\documentclass{arxiv}

\usepackage{url,hyperref,lineno,microtype,subcaption}
\usepackage[onehalfspacing]{setspace}

\usepackage{threeparttable} 
\usepackage{siunitx} 
\usepackage{textcomp}

\title{Enhancing ALS Progression Tracking with Semi-Supervised ALSFRS-R Scores Estimated from Ambient Home Health Monitoring}

\author
{Noah Marchal\,$^{1,3}$\footnote{Correspondence E-mail: noah.marchal@health.missouri.edu},  
William E. Janes\,$^{2}$, 
Mihail Popescu\,$^{1,3}$,
and Xing Song\,$^{1,3}$\\
\vspace{1em} 
\normalfont{\small $^{1}$Institute for Data Science and Informatics;}
\normalfont{\small $^{2}$Department of Occupational Therapy;}
\normalfont{\small $^{3}$Department of Biomedical Informatics, Biostatistics, and Medical Epidemiology; University of Missouri, Columbia, Missouri, United States \vspace{2em} } 
}

\begin{document}

\maketitle
\thispagestyle{firstpagestyle} 

\begin{abstract}
Clinical monitoring of functional decline in ALS relies on periodic assessments that may miss critical changes occurring between visits when timely interventions are most beneficial. 
To address this gap in ALS progression tracking, semi-supervised regression models were developed to estimate rates of decline with pseudo-labeling in a three-patient ALS case series cohort by targeting ALSFRS-R scale trajectories with continuous in-home sensor monitoring data. 
Our analysis compared three model paradigms (individual batch learning and cohort-level batch versus incremental fine-tuned transfer learning) across linear slope, cubic polynomial, and ensembled self-attention pseudo-label interpolations. 
Results revealed cohort homogeneity across functional domains responding to learning methods, with transfer learning improving prediction error for ALSFRS-R subscales in 28 of 32 contrasts (mean RMSE=0.20±0.04), and individual batch learning for predicting the composite scale (mean RMSE=3.15±1.25) in 2 of 3. 
Self-attention interpolation achieved the lowest prediction error for subscale-level models (mean RMSE=0.19±0.06), capturing complex nonlinear progression patterns, outperforming linear and cubic interpolations in 20 of 32 contrasts, though linear interpolation proved more stable in all ALSFRS-R composite scale models (mean RMSE=0.23±0.10).
We identified distinct homogeneity-heterogeneity profiles across functional domains with respiratory and speech functions exhibiting patient-specific patterns benefiting from personalized incremental model adaptation, while swallowing and dressing functions followed cohort-level trajectories suitable for population-based models. 
These findings suggest that dynamically matching learning and pseudo-labeling techniques to functional domain-specific homogeneity-heterogeneity profiles enhances predictive accuracy in ALS progression tracking. 
Integrating adaptive model selection within sensor monitoring platforms could enable timely interventions and scalable deployment in future multi-center studies.
\end{abstract}

\section{Introduction}

ALS is a neurodegenerative disease affecting the motor neuron system with patients experiencing significant difficulties performing across a range of functions leading to reduced ability for self-care. 
Decline in function is measured regularly at provider visits through clinical instruments such as the Revised Amyotrophic Lateral Sclerosis Functional Rating Scale (ALSFRS-R) \citep{Rutkove2015}. 
However, acute functional decline may go undetected by clinicians until the next follow-up due to the length of time between office visits. 
Sensor monitoring, which has been shown to be effective in supporting care for older adults living independently, offers a possible solution for tracking functional changes related to disease progression in those living with ALS. 
Sensor measurements may serve as predictive features to target instrument scales over interim periods between clinic visits, thereby increasing the fidelity of functional measures to aid clinicians in making better, more informed care strategies to guide interventions. 
In this study, we trained and evaluated three semi-supervised learning models (participant-level batch and cohort-level transfer with batch vs. incremental fine-tuning) across three pseudo-label techniques (linear, cubic, and self-attention interpolation) at predicting ALSFRS-R scale trajectories from in-home sensor health features using root mean square error (RMSE) and Pearson's correlation ($r$) as primary metrics of model accuracy and fit. 

\subsection{Sensor Monitoring of ALS Progression}

Sensor-based health monitoring has been shown to improve clinical outcomes in older adult independent living residents through early illness detection, enabling them to maintain their independence longer \citep{Skubic2015}. 
Physical deficits in older adults may mirror the functional declines observed in ALS, with community-dwelling older adults experiencing stable physical function until a steep decline 1–3 years before death \citep{Stolz2024}. 
Additionally, age-related frailty may involve motor unit loss (denervation) similar to ALS, contributing to muscle wasting which could further exacerbate ALS progression in older patients \citep{Verschueren2022}. 
This evidence indicates that remote sensor monitoring technologies effective for improving care in elder populations may identify digital biomarkers for tracking ALS disease progression. 
Recent research has found that combining wearable sensor data with self-reported clinic assessments and environmental metrics improves predictive models targeting ALSFRS-R scales \citep{Marinello2024}. 
Similarly, work evaluating wearable accelerometer, ECG, and digital speech sensors for tracking ALS has shown that changes in physical activity, heart-rate, and speech features correlate with decline in ALSFRS-R scales \citep{Kelly2020}. 
More frequent, remote sensor–based tracking of changes in ALSFRS-R scales would enable clinicians to better target interventions and detect acute events, such as falls or medication changes, between clinic visits. 

\subsection{Clinical Use of ALSFRS-R Scales}

ALS disease progression rates vary between patients by a number of clinical factors including baseline functional status, disease stage at diagnosis and diagnostic delay, co-occurrence of frontotemporal dementia, sex, age and site at onset, particularly respiratory-onset, in addition to a number of genetic and environmental factors \citep{Requardt2021, Su2021, Engelberg2024}. 
Disease progression also varies across functional domains within ALS patients, following nonlinear rates of decline in specific areas \citep{Ramamoorthy2022}. 
ALS progression is tracked longitudinally using the ALSFRS-R instrument as a qualitative, subjective self-reported measure of performance in functional tasks. 
ALSFRS-R scales are collected during clinic visits to determine the amount of change in bulbar, fine motor, gross motor, and respiratory functional domains over time.  
Scales are rated between 0 and 4, with 0 indicating dependence and 4 indicating no difficulty. 
The composite score and linear slope serve as primary metrics of functional change and decline progression and for measuring intervention effects within individuals or across treatment groups in clinical trials, with more frequent assessment improving slope estimation \citep{vanEijk2020, Erb2024}. 
Due to the multi-dimensional aspect the aggregate ALSFRS-R composite score, it has been suggested to use the component scales independently for measuring treatment outcomes \citep{Franchignoni2013}. 
As such, there isn't a one-size-fits all approach for monitoring progression, as decline varies nonlinearly among patients, and individualized  clinical models are needed for tracking across ALSFRS-R functional domains.  

\section{Materials and Methods}

\subsection{Parent Study}
Participants were recruited for a single-site single-cohort prospective study overseen by the MU Institutional Review Board through the MU Health ALS Clinic, with inclusion criteria requiring an ALS diagnosis, residence within 100 miles of the clinic, and either a home caregiver or Montreal Cognitive Assessment (MoCA) score greater than 22, to investigate continuous, in-home sensor monitoring for tracking between-clinic-visit functional decline \citep{Janes2025}. 
ALSFRS-R scores were collected monthly by telephone and quarterly as pre-clinic assessments. 
After accounting for length of enrollment, data from three participants were of sufficient length of at least 6 months for case-series modeling, described in Table \ref{tab:demographics_individual}. 
All three participants were non-Hispanic, white, male, medicare recipients, and left the study due to death. 
The in-home sensor monitoring systems, licensed by the University of Missouri to Foresite Healthcare, LLC., are comprised of three modalities for continuous contactless data collection: bed mattress hydraulic transducers for recording ballistocardiogram (BCG) derived respiration, pulse, and sleep restlessness measures \citep{Heise2010, Rosales2017}; privacy-preserving thermal depth sensors \citep{Stone2013TBME, wallace2017PH}, which detect falls and collect walking speed, stride time, and stride length measurements, though gait data was excluded due to wheelchair use; and passive infrared (PIR) motion sensors that provide room activity counts. 

\subsection{Estimating Between Visit Change in ALSFRS-R Scales}

\begin{figure}[htp]
  \centering
  \begin{subfigure}[t]{\textwidth}
    \includegraphics[width=\linewidth]{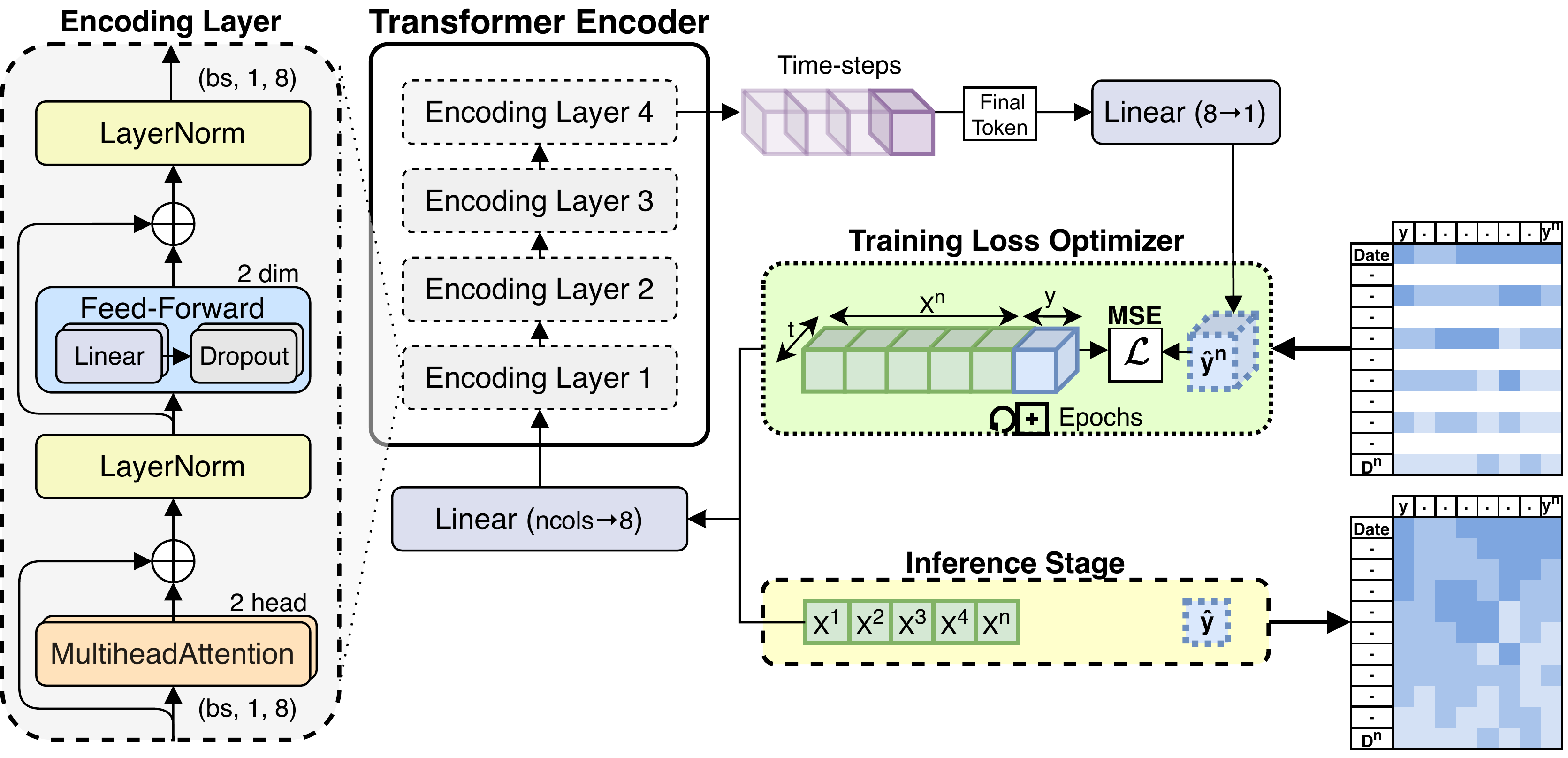}
    \caption{Self-attention interpolation transformer architecture.}
    \label{fig:panel_interpolation_A_transformer}
  \end{subfigure}
  \vspace{1em} 
  \begin{subfigure}[t]{\textwidth}
    \includegraphics[width=\linewidth]{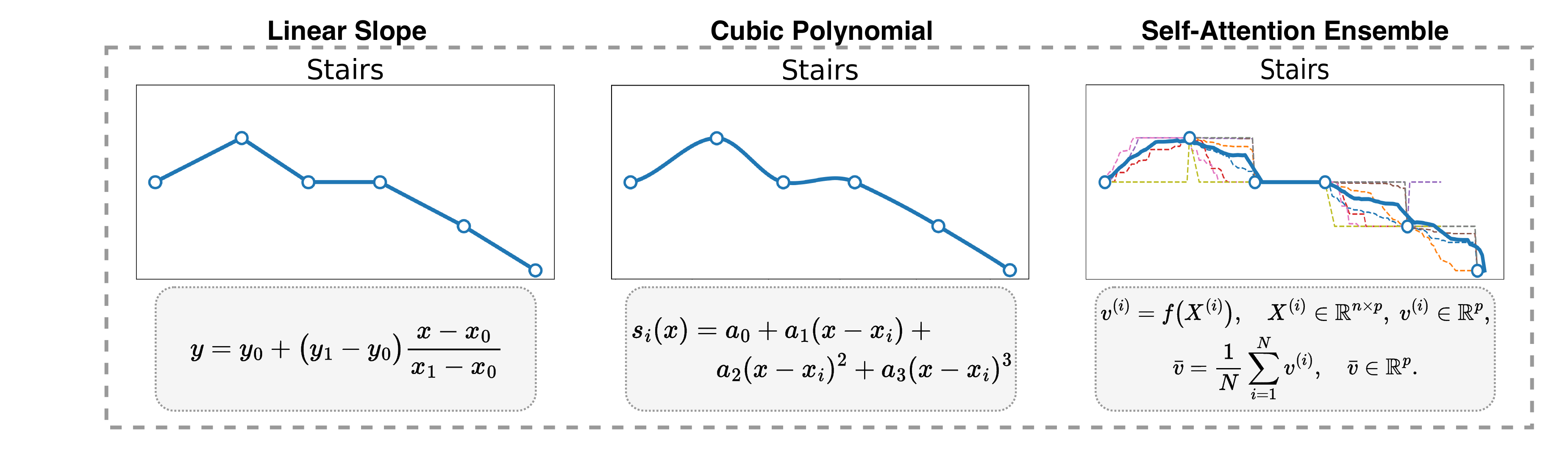}
    \caption{Comparison of interpolation effects.}
    \label{fig:panel_interpolation_B_compare}
  \end{subfigure}
  \caption{Interpolation techniques applied to ALSFRS-R subscores for estimating sensor feature pseudo-labels.}
  \label{fig:panel_interpolation}
\end{figure}

\begin{figure}[htbp]
    \begin{center}
        \includegraphics[width=\textwidth]{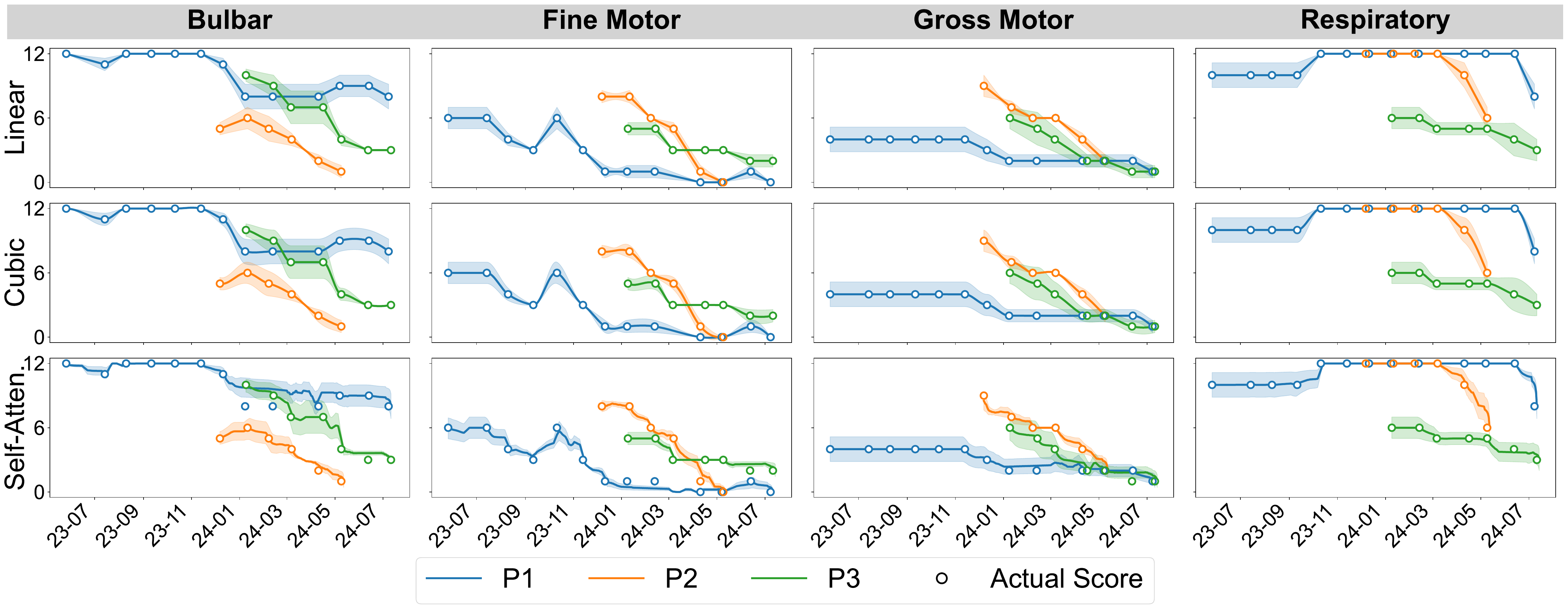}
    \end{center}
    \caption{Participant aggregated ALSFRS-R subscores by functional domain and interpolation technique.}
    \label{fig:interpolate_combined_sum_bar}
\end{figure}

ALSFRS-R scales were aligned at matching frequency to daily aggregated sensor measurements using pseudo-labels for semi-supervised regression, extending prior work evaluating between-visit interpolation \citep{Marchal2024}. 
We incorporated a transformer encoder architecture for self-attention interpolation as compared to polynomial functions, illustrated in Fig. \ref{fig:panel_interpolation_A_transformer}. 
Linear 1-d piecewise interpolation served as a baseline method consistent with clinical methodology for tracking ALS progression.  
Nonlinear cubic spline interpolations were applied to evaluate more gradual rates of decline. 
The transformer encoder mapped date-indexed sensor vectors with known ALSFRS-R ratings to estimate the amount of change occurring between collection points, with the architecture intentionally kept shallow to provide continuous values rather than predicting crisp labels with a deeper network \citep{heaton2022}. 
To further smooth the estimations, self-attention interpolation was applied to each sensor feature algorithm table and average ensembled, shown as dashed plots in Fig. \ref{fig:panel_interpolation_B_compare}. 
The resulting interpolated slopes over time for each pseudo-labeling technique, summated by functional area in Fig. \ref{fig:interpolate_combined_sum_bar}, demonstrate varying rates of decline across participants. 

\subsection{Semi-Supervised Learning of ALSFRS-R Scales}
Three learning approaches for training semi-supervised regression models were compared: batch models fit on sequential individual-level data and transfer models initialized on randomized group-level data and fine-tuned with batch or incremental learning on individual-level data \citep{Hakkal2024}, as shown in Fig. \ref{fig:panel_processing_B_dataframes}. 
Individual batch models were fit on feature sets split proportionally into training and test datasets, while group transfer models employed a leave-one-out design initialized on the remaining participants' observations \citep{Bousquet2004, Torrey2010}. 
Participant-subscales exhibiting very low variance in training samples were not modeled with individual batch learning. 

\subsubsection{Data Preprocessing and Feature Engineering}

The preprocessing pipeline, described in Fig. \ref{fig:panel_processing_A_preproc}, began by segmenting time-indexed features into day and night periods. 
Summary statistics were calculated over each feature channel and period for count, minimum, maximum, mean, median, mode, variance, range, skew, kurtosis, quantiles, inter-quartile range (IQR), coefficient of variation (CV), and entropy to better capture temporal patterns by time-of-day. 
High collinear features were then removed and the resulting feature set was normalized to its minimum-maximum. 
ALSFRS‑R scores were interpolated with each technique, resulting in three continuous target trajectories per participant. 
Finally, the normalized features were pivoted by period and joined by date with each target interpolation to produce pseudo-labeled datasets. 

\subsubsection{Iterative Hyperparameter Tuning and Feature Selection}

\begin{figure}[htp]
  \centering
  \begin{subfigure}[t]{\textwidth}
    \includegraphics[width=\linewidth]{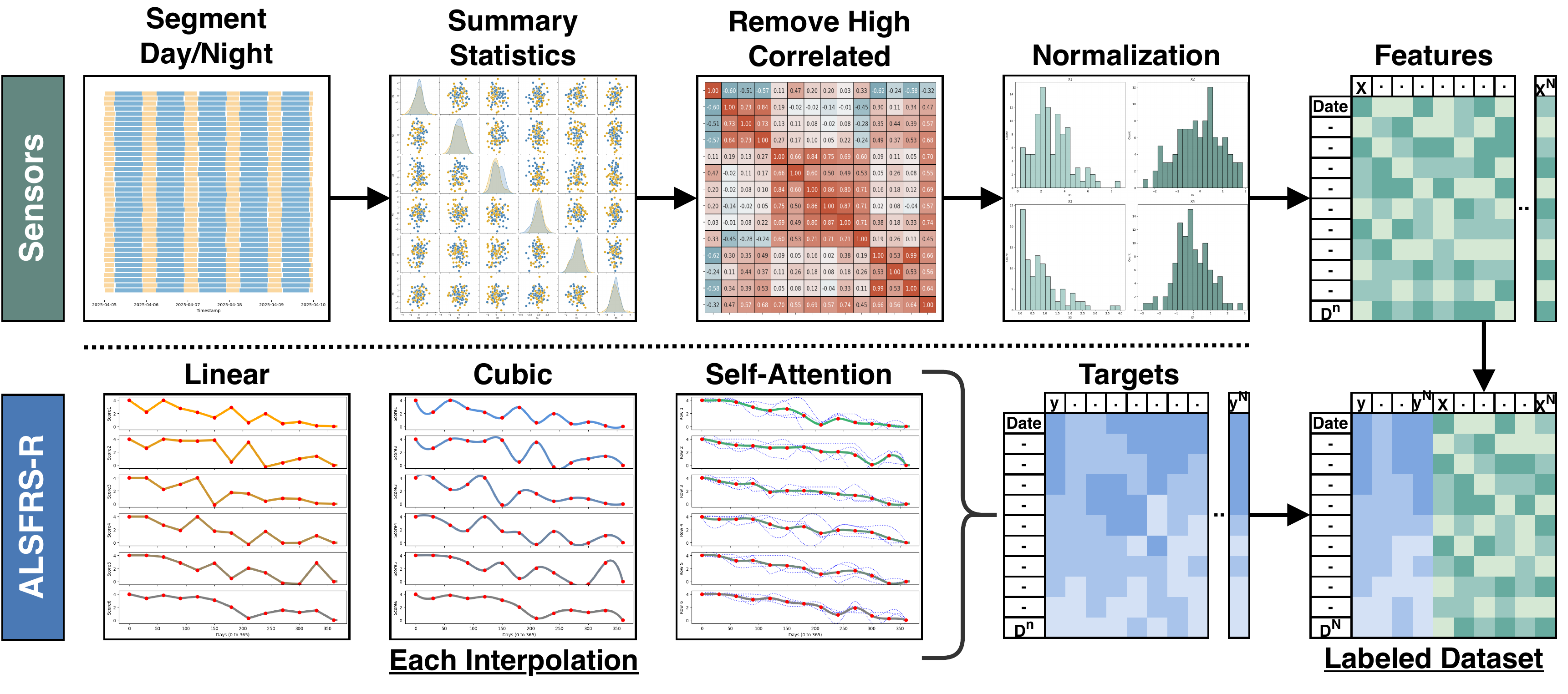}
    \caption{Sensor and ALSFRS-R preprocessing steps.}
    \label{fig:panel_processing_A_preproc}
  \end{subfigure}
  \vspace{1em} 
  \begin{subfigure}[t]{\textwidth}
    \includegraphics[width=\linewidth]{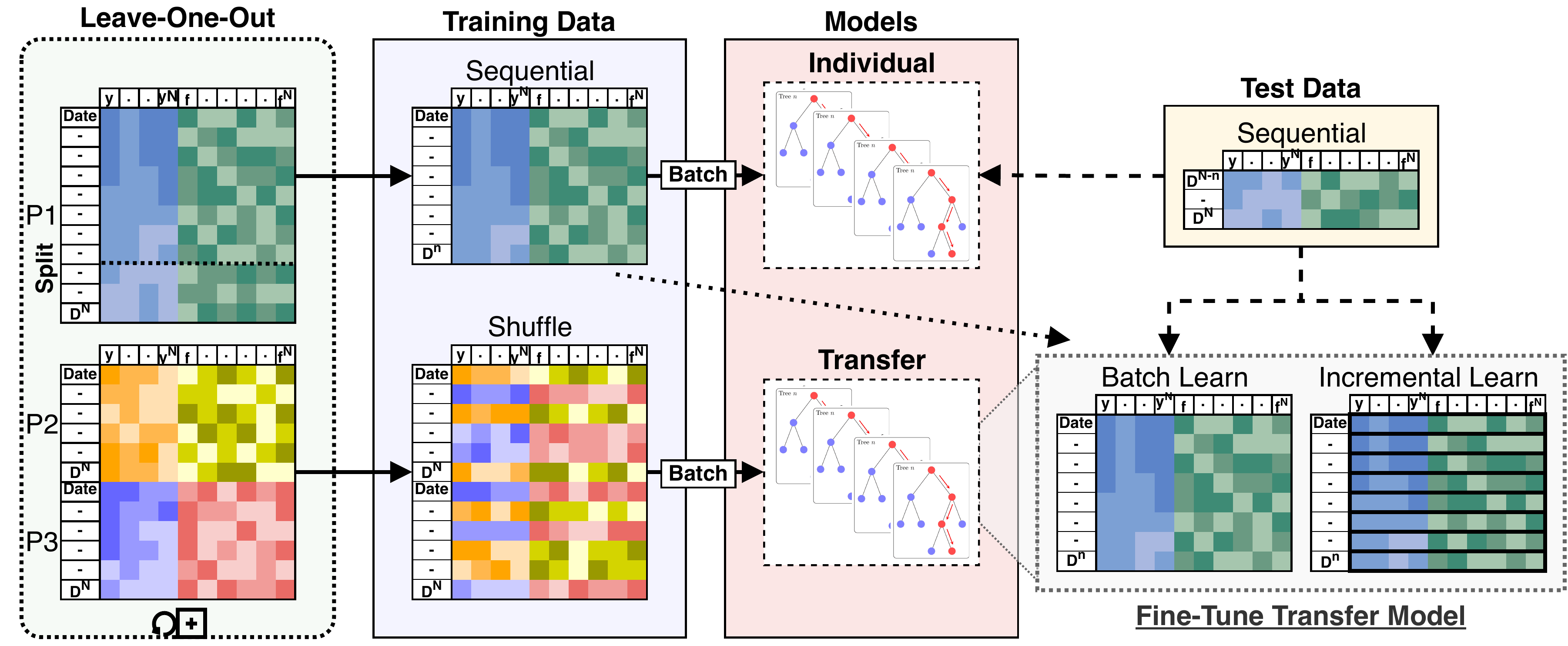}
    \caption{Dataset segmentation for individual batch and cohort transfer learning models.}
    \label{fig:panel_processing_B_dataframes}
  \end{subfigure}
  \vspace{1em} 
  \begin{subfigure}[t]{\textwidth}
    \includegraphics[width=\linewidth]{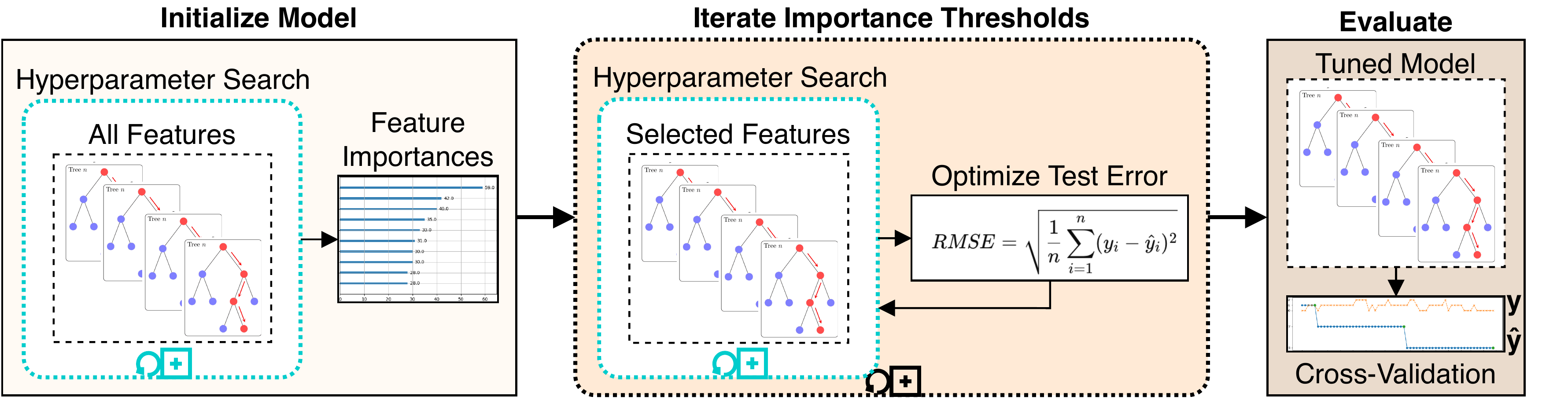}
    \caption{Iterative model tuning and evaluation stages.}
    \label{fig:panel_processing_C_models}
  \end{subfigure}
  \caption{Data processing and model fitting pipeline.}
  \label{fig:panel_processing}
\end{figure}

We employed an iterative screener-learner approach combining hyperparameter optimization with feature selection, illustrated in Fig. \ref{fig:panel_processing_C_models}, using the XGboost algorithm \citep{Chen2016}. 
Models were initialized on the full feature set and hyperparameter tuned using RandomizedSearchCV in the Scikit-Learn Python package. 
Following initialization, feature importance scores were extracted using feature 'weight' frequency internal to XGBoost. 
Importance scores were then binned through iterative precision-rounding, beginning at 6 decimal places and consecutively increasing precision levels to a maximum of 200 features to avoid overfitting. 
For each feature subset identified, a new model was fit using the selected features and tuned hyperparameters, as specified in Table \ref{tab:hyperparameters}. 
Root mean square error (RMSE) was calculated and compared across all iterations, with the model configuration yielding the lowest test RMSE selected as the optimized model. 

\subsection{Model Evaluation Metrics and Taylor Diagrams}

Model performances were evaluated using RMSE, Pearson correlation ($r$), and standard deviation (SD) for modeled outcomes ($\hat{y}$) against test values ($y$) held-out from model training data. 
Comparisons across pseudo-labeling techniques and learning methods were made using Taylor diagrams to assess prediction accuracy, correlation, and variability at the cohort level by averaging outcomes across participants for each subscale. 
In Taylor diagrams, the reference point on the X-axis represents actual values (estimated ALSFRS-R scale) plotted at coordinates \((\sigma_{\text{ref}}, 0)\), where \(\sigma_{\text{ref}}\) is the standard deviation of the actual values, while the origin \((0,0)\) represents a modeled prediction with SD=0 and $r$=0 to the reference vector \citep{Taylor2001}. 

\section{Results}

Models were evaluated across pseudo-labeling techniques (linear, cubic, self-attention interpolation) and learning method (individual batch, transfer batch fine-tuned, transfer incremental fine-tuned) for predicting ALSFRS-R component and composite scales with in-home health sensor features. 
ALSFRS-R subscale prediction errors and outcome correlations are illustrated as Taylor diagrams by participant in Fig. \ref{fig:TD_all_userids_nonorm} and cohort average in Fig. \ref{fig:TD_avg_userids_nonorm}. 
Low variances in collected scores, provided in Table \ref{tab:variance_scores}, prevented fitting participant-subscale models for Salivation in P1, and Speech in P2, and Dyspnea and Respiratory for P1 and P2 for all interpolations, and for Dyspnea and Respiratory in P3 with linear and cubic interpolation, and Cutting with self-attention interpolation. 
As P1 Salivation and Respiratory scales had zero variance with static scores of 4, transfer models were not fit for those subscales. 

\subsection{Performance Contrast Between Learning Methods} 

In this section, individual batch and cohort-level transfer learning methods are compared for improving prediction error and outcome correlation within participants as detailed in Table \ref{tab:result_learning_alsfrs}, and as averaged across participants in Table \ref{tab:result_learning_avg_alsfrs}. 
Results from individual batch and transfer incremental models reveal participant-specific heterogeneity in certain subscales. 
Comparing the lowest and next lowest model errors, transfer learning generally outperforms individual batch learning across most participant-scale combinations, with exceptions (P1:Walking, Stairs, Composite; P2:Composite; P3:Cutting, Turning, Orthopnea) where individual batch models provided lower prediction error.

\subsubsection{Bulbar Function} 

The bulbar ALSFRS-R scales (Speech, Salivation, Swallowing) demonstrated cohort-level trends with both transfer learning models outperforming individual batch models. 
For Speech, improvements in mean prediction error and correlation from transfer incremental learning were observed in P1 (RMSE=0.08$\rightarrow$0.04, $r$=–0.02$\rightarrow$0.22) and P2 (RMSE=0.27$\rightarrow$0.26, $r$=0.56$\rightarrow$0.75) compared to transfer batch, while P3 had an increase in mean prediction error and little change in correlation (RMSE=0.15$\rightarrow$0.19, $r$$\sim$0), indicating that Speech function has shared cohort-level substrates with participant-specific changes captured better by transfer incremental fine-tuning over individual or transfer batch models. 
For Salivation, transfer incremental resulted in negligible change in mean prediction error in P2 (RMSE=0.21$\rightarrow$0.20) while error increased in P3 (RMSE=0.15$\rightarrow$0.34) compared to transfer batch learning. 
Both participants gained minor prediction correlation (P2 $r$=-0.28$\rightarrow$-0.20) (P3 $r$=0.0$\rightarrow$0.03), suggesting slight participant heterogeneity in Salivation decline. 
For Swallowing, transfer incremental tuning increased prediction error with decreased correlation in P1 (RMSE=0.21$\rightarrow$0.23, $r$=0.80$\rightarrow$0.62) and with no change in correlation for P3 (RMSE=0.10$\rightarrow$0.16, $r$=0.0). 
There was no change in P2 error or correlation (RMSE=0.04, $r$=0.0). 

\subsubsection{Fine Motor Function} 

Fine-motor subscales (Handwriting, Cutting, and Dressing) had variable responses to individual batch and transfer incremental fine-tuning across participants with overall weak participant heterogeneity. 
Individual batch models were outperformed by transfer learning models implying that fine‐motor decline progression patterns are relatively homogeneous to the cohort. 
For Handwriting, transfer incremental fine‐tuning improved mean prediction error and correlation for P1 (RMSE=0.35$\rightarrow$0.15, $r$=–0.02$\rightarrow$0.08) and for P2 (RMSE=0.12$\rightarrow$0.10, $r$=0.32$\rightarrow$0.46), whereas P3 saw a slight increase in prediction error (RMSE=0.06$\rightarrow$0.10) with no change in correlation ($r$=0.0). 
In contrast, the Cutting subscale exhibited more cohort homogeneity as only P2 improved with transfer incremental learning (RMSE=0.08$\rightarrow$0.05, $r$=–0.22$\rightarrow$0.28), while P1 (RMSE=0.09$\rightarrow$0.12) and P3 (RMSE=0.02$\rightarrow$0.03) had increased prediction error without correlation improvement ($r$=0.0) compared to transfer batch. 
Dressing subscale heterogeneity was more moderate between participant and cohort with slight improvements from transfer incremental learning in mean RMSE and $r$ for P2 (RMSE=0.11$\rightarrow$0.06, $r$=0.94$\rightarrow$0.99) and P3 (RMSE=0.16$\rightarrow$0.15, $r$=0.31$\rightarrow$0.37) while both error and correlation worsened in P1 (RMSE=0.27$\rightarrow$0.35, $r$=0.40$\rightarrow$-0.20). 

\subsubsection{Gross Motor Function} 

For gross motor ALSFRS-R scales (Turning, Walking, Stairs) individual batch models outperformed transfer batch models in a few examples indicating heterogeneity in those cases. 
Individual batch learning demonstrated lower error with a minor correlation decrease in the Turning subscale for P3 (RMSE=0.15$\rightarrow$0.16, $r$=–0.08$\rightarrow$0.08) compared to transfer batch. 
Similarly, individual batch had better model error with lower correlation than transfer batch learning in the Walking subscale for P1 (RMSE=0.05$\rightarrow$0.25, $r$=0.10$\rightarrow$0.24) and marginally in Stairs for P1 (RMSE=0.0$\rightarrow$0.04, $r$=0). 
For Turning, transfer incremental improved both error and correlation in P1 (RMSE=0.32$\rightarrow$0.26, $r$=0.09$\rightarrow$-0.03) and increased error but also correlation in P2 (RMSE=0.03$\rightarrow$0.12, $r$=0.24$\rightarrow$0.67) compared to transfer batch. 
Walking subscale performance also had mixed results between transfer methods. 
Transfer incremental learning improved mean error and correlation over transfer batch for P2 (RMSE=0.14$\rightarrow$0.06, $r$=0.93$\rightarrow$0.98) and only slightly improved P3 (RMSE=0.17$\rightarrow$0.16, $r$=–0.38$\rightarrow$–0.37), but with minor error increase and no change in correlation for P1 (RMSE=0.25$\rightarrow$0.26, $r$=0.24). 
For Stairs, transfer batch and incremental models had marginal difference in error and increased correlation for P2 (RMSE=0.24$\rightarrow$0.23, $r$=0.88$\rightarrow$0.92), while model error increased with transfer incremental learning in P3 (RMSE=0.03$\rightarrow$0.19, $r$=0). 

\subsubsection{Respiratory Function} 

Within the respiration related ALSFRS-R scales (Dyspnea, Orthopnea, Respiratory), the collected Orthopnea and Respiratory scores for P1 and P2 exhibited low variance, shown in Table \ref{tab:variance_scores}, which prevented individual batch model fitting for those participants. 
For Dyspnea, transfer incremental learning significantly improved correlation for all participants and reduced prediction error for P1 (RMSE=0.50$\rightarrow$0.42, $r$=0.23$\rightarrow$0.75) but increased error for P2 (RMSE=0.46$\rightarrow$0.54, $r$=0.88$\rightarrow$0.94) and P3 (RMSE=0.14$\rightarrow$0.36, $r$=–0.03$\rightarrow$0.47). 
In Orthopnea, transfer incremental models resulted in marginal differences in prediction error and slight improvement in correlation for P1 (RMSE=0.12$\rightarrow$0.11, $r$=0.98$\rightarrow$0.99) and P2 (RMSE=0.31$\rightarrow$0.30, $r$=0.67$\rightarrow$0.73). 
However, individual batch models had slightly better error but lower correlation compared to transfer batch in P3 (RMSE=0.10$\rightarrow$0.12, $r$=0.39$\rightarrow$0.60), indicating that the P3 Orthopnea trajectory wasn't fully captured by the shared model, pointing towards participant heterogeneity given the small improvements in P1 and P2 and benefit in P3 from individual batch learning. 
Similar to orthopnea, respiratory function displayed participant heterogeneity with transfer incremental learning resulting in slight change in error but with an increase in correlation between transfer batch and incremental learning for P2 (RMSE=0.30$\rightarrow$0.31, $r$=0.28$\rightarrow$0.48), while P3 had improved error and significant increase in correlation (RMSE=0.16$\rightarrow$0.12, $r$=–0.27$\rightarrow$0.79). 

\subsubsection{Composite ALSFRS-R Scale}

As a summary of overall functional status across bulbar, fine-motor, gross-motor and respiratory items, the composite ALSFRS-R scale captures decline trends as a single index. 
For the composite ALSFRS-R scale, individual batch learning models resulted in lower mean prediction error compared to transfer batch learning with marginal difference in outcome correlation for P1 (RMSE=2.76$\rightarrow$3.29, $r$=0.03$\rightarrow$-0.04) and with significantly better error and correlation in P2 (RMSE=3.52$\rightarrow$6.03, $r$=0.48$\rightarrow$0.11), demonstrating that modeled composite scores are mostly participant-specific. 
Transfer incremental learning slightly improved prediction error and correlation in P3 (RMSE=3.01$\rightarrow$2.97, $r$=-0.54$\rightarrow$-0.41) over transfer batch models, compared to individual batch learning error (RMSE=3.17, $r$=-0.18). 

\subsection{Performance Contrast Between Pseudo-Label Interpolations}

\begin{figure}[htbp]
    \centering
    \includegraphics[width=\textwidth]{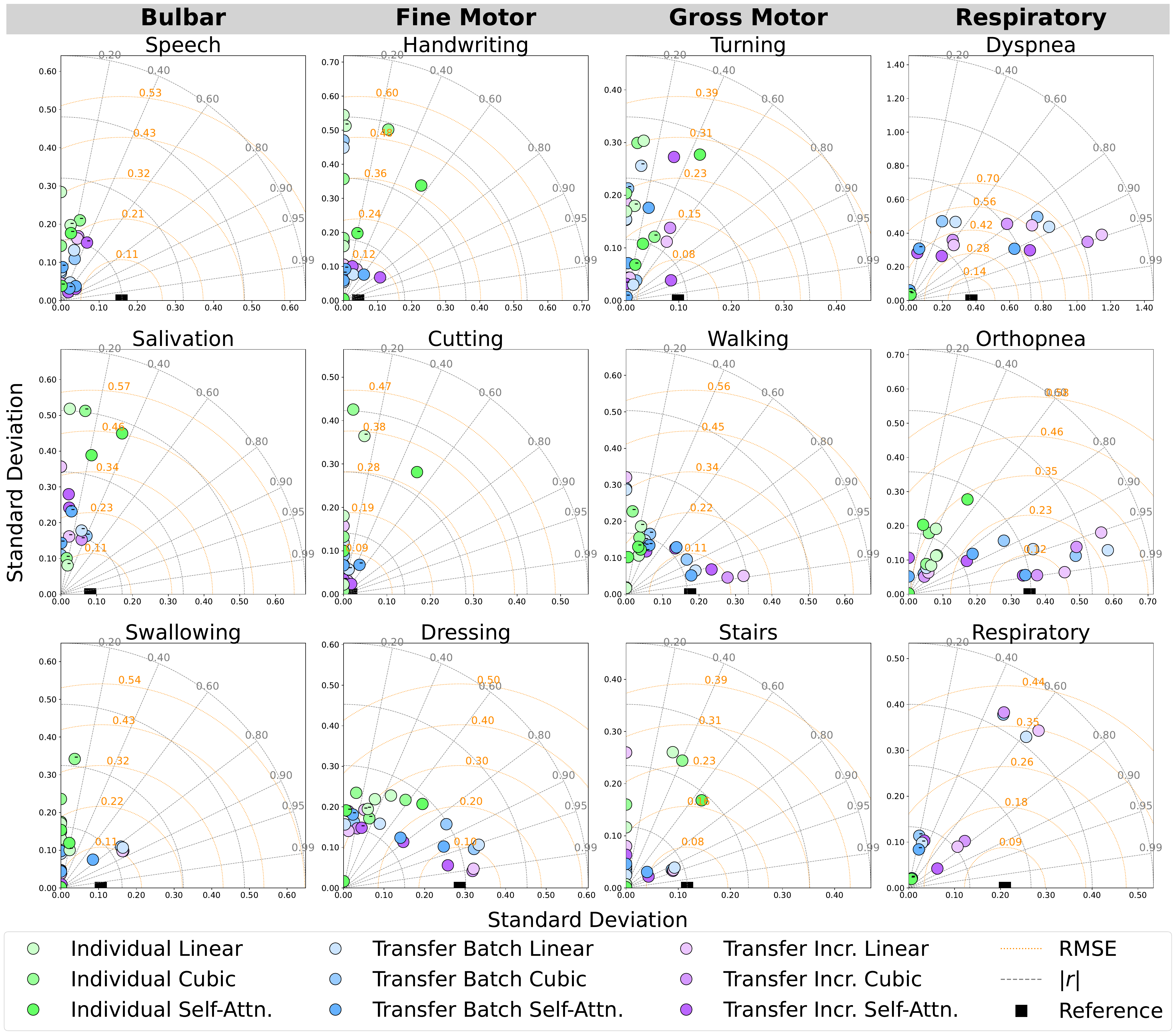}
    \caption{Participant-level Taylor diagrams depicting mean RMSE, absolute correlation ($|r|$), and standard deviation of predicted outcomes for each ALSFRS-R scale, annotated by negative ($-$) correlation.}
    \label{fig:TD_all_userids_nonorm}
\end{figure}

To evaluate how the pseudo-labeling interpolation approach affects model performance, prediction error and outcome correlation were averaged across learning methods as detailed in Table \ref{tab:result_interpolation_alsfrs} and across participants in Table \ref{tab:result_interpolation_avg_alsfrs}. 
Results demonstrate that non-linear cubic polynomial and self-attention interpolation of ALSFRS-R scales follow more closely with daily changes in in-home sensor health measurements, with a few exceptions (P1:Composite; P2:Dressing, Orthopnea, Composite; P3:Turning, Respiratory, Composite) where linear interpolation resulted in lower error, illustrated in Taylor diagrams in Fig. \ref{fig:TD_all_userids_nonorm}. 

\subsubsection{Bulbar Function} 

The bulbar ALSFRS-R scales (Speech, Salivation, Swallowing) resulted in mean improvements to model error from non-linear cubic and self-attention interpolation compared to linear. 
For Speech, cubic reduced model error in P1 (RMSE=0.14$\rightarrow$0.09, $r$=0.0) while self-attention provided improved error and correlation in P2 (RMSE=0.30$\rightarrow$0.20, $r$=0.59$\rightarrow$0.75) and improved error at decreased correlation in P3 (RMSE=0.25$\rightarrow$0.20, $r$=0.13$\rightarrow$-0.34). 
Salivation performed only marginally better with cubic interpolation over linear with a decrease in correlation in P2 (RMSE=0.29$\rightarrow$0.28, $r$=-0.13$\rightarrow$-0.29) and minor correlation change in P3 (RMSE=0.20$\rightarrow$0.19, $r$=-0.07$\rightarrow$-0.05). 
For Swallowing scales, self-attention interpolation provided lower error in all participants with decreased correlation in P1 (RMSE=0.27$\rightarrow$0.16, $r$=0.49$\rightarrow$0.36) and no change in correlation for P3 (RMSE=0.12$\rightarrow$0.09, $r$=0.0), while cubic and self-attention performed the same with a very minor difference in error for P2 (RMSE=0.09$\rightarrow$0.08, $r$=0.0) against linear. 

\subsubsection{Fine Motor Function} 

In fine-motor subscales (Handwriting, Cutting, and Dressing), self-attention interpolation provided the lowest model error with the exception of Cutting for P3 and Dressing for P2. 
For the Handwriting scale, self-attention reduced error with increased or no change in correlation for P1 (RMSE=0.39$\rightarrow$0.14, $r$=0.0), P2 (RMSE=0.23$\rightarrow$0.20, $r$=0.24$\rightarrow$0.67), and P3 (RMSE=0.11$\rightarrow$0.05, $r$=0.0). 
Cutting also performed better on self-attention pseudo-labels with reduced error in P1 (RMSE=0.15$\rightarrow$0.07, $r$=0.0) and for P2 (RMSE=0.15$\rightarrow$0.14, $r$=-0.01$\rightarrow$0.21) with improved correlation, while P3 cubic and linear models performed equally (RMSE=0.02, $r$=0). 
Dressing models fit on self-attention interpolated labels had decreased model error but with lower correlation in P1 (RMSE=0.32$\rightarrow$0.27, $r$=0.19$\rightarrow$0.09) and with significantly better correlation in P3 (RMSE=0.25$\rightarrow$0.14, $r$=-0.06$\rightarrow$0.52), while linear slope provided best error and correlation compared to non-linear interpolations in P2 (RMSE=0.16, $r$=0.80). 

\subsubsection{Gross Motor Function} 

For gross motor ALSFRS-R scales (Turning, Walking, Stairs), self-attention interpolation again resulted in improved model error in most cases. 
The Turning subscale had mixed outcomes for model error with cubic improving over linear but with decreased correlation for P1 (RMSE=0.31$\rightarrow$0.26, $r$=-0.01$\rightarrow$-0.10) and with self-attention improving error and correlation for P2 (RMSE=0.15$\rightarrow$0.13, $r$=0.36$\rightarrow$0.41) and minor change in error at improved correlation for P3 (RMSE=0.17$\rightarrow$0.18, $r$=0$\rightarrow$0.10). 
For Walking, self-attention interpolation resulted in the lowest error for all participants with improved correlation in P1 (RMSE=0.21$\rightarrow$0.14, $r$=0$\rightarrow$0.59) and moderate decrease in correlation for P2 (RMSE=0.17$\rightarrow$0.14, $r$=0.75$\rightarrow$0.66) and P3 (RMSE=0.24$\rightarrow$0.21, $r$=-0.29$\rightarrow$-0.37). 
The Stairs models had lowest prediction error with self-attention interpolation with a slight improvement in correlation in P2 (RMSE=0.28$\rightarrow$0.22, $r$=0.73$\rightarrow$0.78) and no change in P3 (RMSE=0.13$\rightarrow$0.04, $r$=0.0).
However, linear slope and cubic interpolation performed equally for P1 (RMSE=0.04, $r$=0.0). 

\subsubsection{Respiratory Function} 

Within the respiration related ALSFRS-R scales (Dyspnea, Orthopnea, Respiratory), more models fit on linear slope demonstrated better error over non-linear interpolations among all functional domains. 
For the Dypsnea scale, self-attention interpolation had improved error with decreased correlation in P1 (RMSE=0.52$\rightarrow$0.36, $r$=0.68$\rightarrow$0.20) and marginal correlation change in P2 (RMSE=0.55$\rightarrow$0.41, $r$=0.92$\rightarrow$0.91). 
Self-attention interpolation had equal prediction error to baseline linear slope but at decreased correlation in P3 (RMSE=0.24, $r$=0.35$\rightarrow$-0.07). 
For Orthopnea models, transfer batch interpolation improved error with a slight reduction in correlation for P1 (RMSE=0.26$\rightarrow$0.22, $r$=0.79$\rightarrow$0.76) compared to linear slope. 
Self-attention interpolation increased model error over linear but with improved correlation in P2 (RMSE=0.29$\rightarrow$0.36, $r$=0.62$\rightarrow$0.75), while the opposite occurred in P3 (RMSE=0.20$\rightarrow$0.05, $r$=0.82$\rightarrow$0.0) reducing model error and significantly decreasing correlation in P3. 
Respiratory had the lowest error from self-attention but with a significant reduction in correlation in P2 (RMSE=0.34$\rightarrow$0.19, $r$=0.62$\rightarrow$0.02), while linear slope provided a slight decrease in both error and correlation compared to cubic interpolation in P3 (RMSE=0.10$\rightarrow$0.12, $r$=0.24$\rightarrow$0.29). 

\subsubsection{Composite ALSFRS-R Scale} 

For the composite ALSFRS-R scale, linear slope provided the best prediction error but at decreased outcome correlation compared to self-attention interpolation for P1 (RMSE=2.79$\rightarrow$4.11, $r$=0.08$\rightarrow$0.50), P2 (RMSE=4.21$\rightarrow$5.54, $r$=0.39$\rightarrow$0.41), and P3 (RMSE=2.38$\rightarrow$4.24, $r$=-0.52$\rightarrow$-0.15). 
The results suggest that while self-attention was able to best capture changes overall within individual ALSFRS-R subscales, aggregation compensates across those changes leading to a linear trajectory in the ALSFRS-R composite scale confirming to clinical practice. 

\begin{figure}[htbp]
    \centering
    \includegraphics[width=\textwidth]{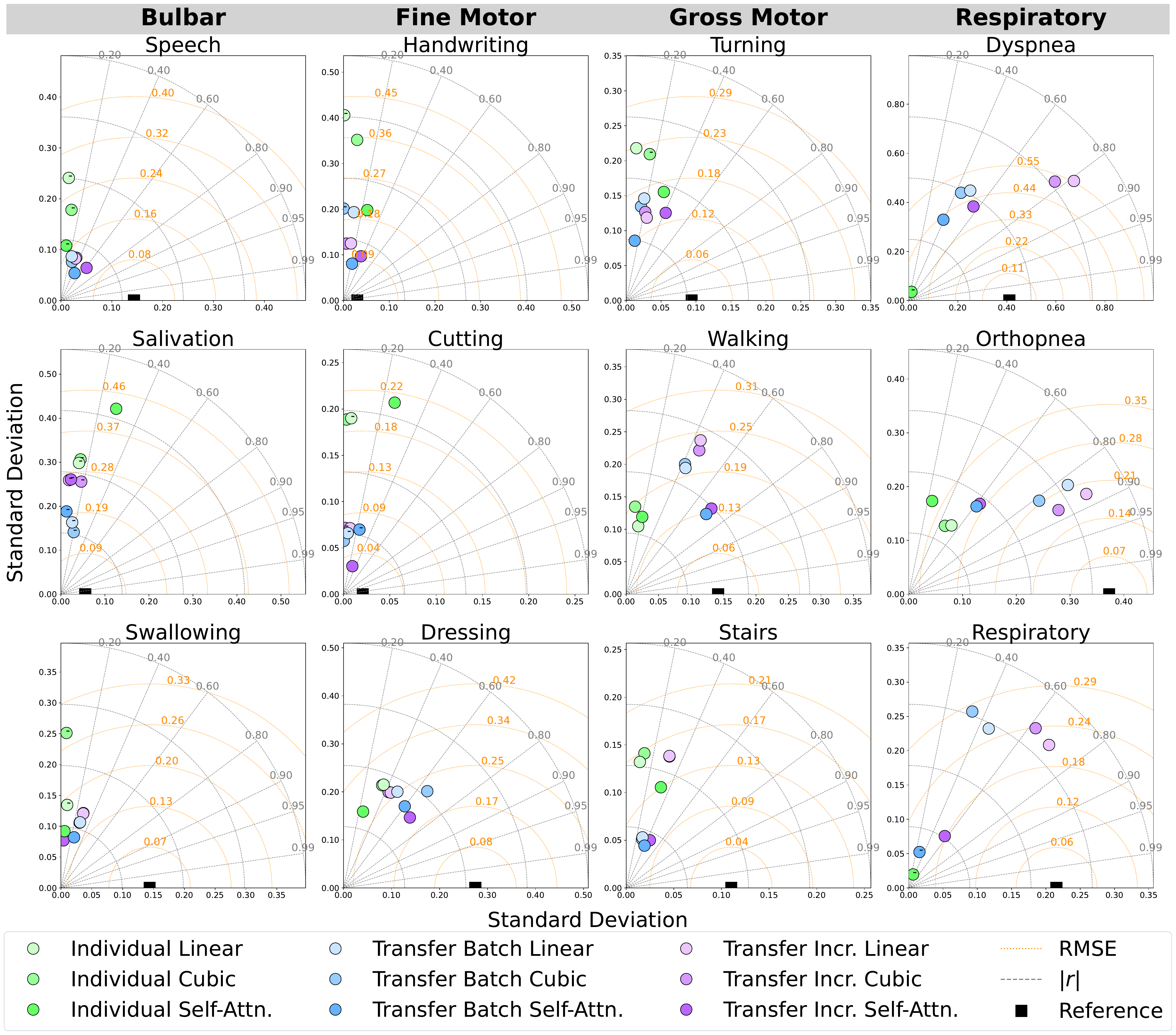}
    \caption{Cohort-averaged Taylor diagrams depicting mean RMSE, absolute correlation ($|r|$), and standard deviation of predicted outcomes for each ALSFRS-R scale, annotated by negative ($-$) correlation.}
    \label{fig:TD_avg_userids_nonorm}
\end{figure}

\section{Discussion}

Semi-supervised learning approaches were evaluated for predicting ALSFRS-R scale trajectories with participant-level batch learning and cohort-initialized transfer learning with batch and incremental fine-tuning strategies from in-home sensor health features. 
Our findings highlight the benefit of incrementally fine-tuning cohort transfer learning models with individual-level data for optimizing prediction error (RMSE) and outcome correlation (Pearson's $r$). 

\subsection{ALSFRS-R Scales Exhibit Mixed Participant-Cohort Homogeneity} 

ALS decline progression varies across ALSFRS-R functional areas, creating multi-dimensional trajectories where some subscales decline predictably across the cohort while others follow patient-specific trends. 
As illustrated in Fig. \ref{fig:interpolate_combined_sum_bar}, rates of decline in bulbar, gross, and respiratory area measures for P1 were marked by periods of stability followed by sudden decreases compared to the regular decline observed in P2 and P3. 
Similarly, P1 fine motor measures increased near Nov. 2023 followed by a regular rate of decline as with decreases observed P2 and P3. 
Contrasting mean model performances across participants and learning methods provides evidence of whether functional domains exhibit patient-specific heterogeneity or cohort-level homogeneity, reported in Table \ref{tab:result_learning_avg_alsfrs}. 
Bulbar area functions demonstrated mixed patterns with Speech exhibiting a combination of cohort homogeneity and participant heterogeneity with transfer incremental improving outcome error and correlation over individual batch models (RMSE$\approx$0.29$\rightarrow$0.16, $r$$\approx$-0.09$\rightarrow$0.34), while Salivation showed near zero correlation for individual model and persistent negative correlations for transfer models suggesting high noise or complex participant-level patterns, and Swallowing having more cohort homogeneity with transfer batch (RMSE$\approx$0.12, $r$$\approx$0.27) outperforming both individual batch and incremental fine-tuned models. 
Fine motor area functions similarly demonstrated mixed results as Handwriting exhibited cohort patterns benefiting from transfer learning but also participant heterogeneity with incremental fine-tuning improving error and correlation over individual batch (RMSE$\approx$0.31$\rightarrow$0.12, $r$$\approx$0.07$\rightarrow$0.18), Cutting displaying similar heterogeneity with incremental fine-tuning improving correlation performance over transfer batch ($r$=-0.07$\rightarrow$0.09) with near equal prediction error, and Dressing having moderate cohort homogeneity with transfer batch having stronger correlation compared to incremental fine-tuning ($r$$\approx$0.55$\rightarrow$0.39) with no change in error. 
Gross motor area functions were predominantly cohort homogeneous with selective participant heterogeneity, particularly in Walking with incremental fine-tuning marginally improving error and correlation over batch fine-tuning (RMSE$\approx$0.19$\rightarrow$0.16, $r$$\approx$0.26$\rightarrow$0.28) and over individual batch models in Stairs (RMSE$\approx$0.13$\rightarrow$0.17, $r$$\approx$0.15$\rightarrow$0.31) indicating participant variations exist within predominant cohort patterns. 
Respiratory functions exhibited the strongest evidence of patient heterogeneity, with incremental fine-tuning of transfer models improving negative correlations from individual batch in Dyspnea ($r$$\approx$-0.30$\rightarrow$0.72), Orthopnea ($r$$\approx$0.41$\rightarrow$0.79), and Respiratory ($r$$\approx$-0.31$\rightarrow$0.64). 
These patterns suggest tailoring learning methods to the underlying homogeneity-heterogeneity profile of each functional domain might improve model optimization, with cohort homogeneous scales making use of transfer learning, participant heterogeneous scales requiring incremental adaptation to better capture patient-specific patterns, and mixed pattern functions potentially benefiting from adaptive learning approaches when participant-level patterns deviate from cohort-baseline variability. 

\subsection{Transfer Learning Improves Performance in Subscale Models} 

Incremental fine-tuning of transfer learning models provided the best balance between predictive accuracy and correlation when contrasting the results by learning technique. 
The Taylor diagram analysis in Fig. \ref{fig:TD_avg_userids_nonorm} of model predictions aggregated across participants illustrated how transfer learning models outperformed participant-level batch learning across most subscales, aligning predictions closer to reference vectors with improved accuracy and correlation. 
Performance differences between individual batch and transfer learning models were particularly pronounced across subscales, shown in Table \ref{tab:result_learning_alsfrs}, where transfer batch models reduced mean prediction error in 12 of 34 comparisons and transfer incremental learning in 16 of 34, excluding ties with $r$=0. 
Individual batch models exhibited increased error and weaker correlations, achieving best performance in only 4 of 30 ALSFRS-R scale models (P1:Walking; P3:Turning, Orthopnea). 
Across participants and pseudo-labeling techniques, transfer batch models achieved lower mean prediction error and improved correlation (RMSE$\approx$0.18(0.04)) compared to transfer incremental (RMSE$\approx$0.20(0.04)) and individual batch learning (RMSE$\approx$0.25(0.07)), reported in the Subscale mean row of Table \ref{tab:result_learning_avg_alsfrs}. 
Additionally, evaluating outcome correlation, transfer incremental learning increased correlation in 22 of 34 comparisons with higher mean correlation ($r$$\approx$0.35(0.15)) across interpolation techniques and participants compared to transfer batch ($r$$\approx$0.22(0.15)) and individual batch ($r$$\approx$0.02(0.12)) learning methods. 
As such, despite slightly higher mean prediction error, transfer incremental captured individual trajectory patterns more effectively, suggesting better detection of temporal changes in ALS decline. 
Individual batch models demonstrated improved error for composite scales (RMSE$\approx$3.15(1.25), $r$$\approx$0.11(0.10)) compared to transfer batch (RMSE$\approx$4.11(1.19), $r$$\approx$–0.16(0.31)). 
However, transfer incremental models provided the most balanced composite performance (RMSE$\approx$3.54(0.63), $r$$\approx$0.26(0.20)) having better correlation at a slight increase in error. 
The improved performance of transfer learning approaches across subscales suggests ALS progression follows cohort-level patterns predictive of individual trajectories despite disease heterogeneity, supporting that there are underlying shared mechanisms that can be captured through cohort-level models. 
The effectiveness of incremental fine-tuning indicates that personalized ALS progression tracking should incorporate both population parameters and adaptive learning to individual patients. 

\subsection{Selection of Pseudo-labeling Technique is Clinical Task Specific} 

Examining modeled outcomes by pseudo-labeling technique showed that the choice of optimal interpolation is largely outcome and metric specific, with performance varying across subscale and composite measures. 
Self-attention interpolation achieved the best subscale-specific performance with the lowest mean prediction error (RMSE$\approx$0.19(0.06), $r$$\approx$0.24(0.27)) across participants and ALSFRS-R subscales as shown in mean row of Table \ref{tab:result_interpolation_avg_alsfrs}, outperforming linear and cubic interpolation in 20 of 32 subscale comparisons excluding ties, reported in bold in Table \ref{tab:result_interpolation_alsfrs}. 
Linear and cubic interpolations showed nearly identical mean performance (RMSE$\approx$0.23(0.10), $r$$\approx$0.25(0.18)-0.24(0.19)) and were optimal in only 4 and 5 subscale models, respectively. 
This pattern reversed for composite scales, where linear interpolation demonstrated lower error in 2 of 3 comparisons (RMSE$\approx$3.13(0.69)) and cubic in 1 of 3 (RMSE$\approx$3.04(0.69)), while self-attention resulted in increased error but with improved correlation (RMSE$\approx$4.63(1.19)), suggesting that composite trajectories benefit more from stable estimation using linear slope than potentially over-responsive self-attention interpolation. 
However, choice of evaluation metric also factors into pseudo-labeling selection, with self-attention providing the best correlation for composite models when prioritizing prediction-outcome trend agreement ($r$$\approx$0.25(0.38)) compared to linear ($r$$\approx$-0.02(0.28)) and cubic ($r$$\approx$0.02(0.31)). 
These findings indicate that subscale-level monitoring benefits from self-attention pseudo-labeling, composite progression scoring from linear interpolation, and correlation-focused applications from self-attention for composite scales, highlighting the importance of matching analytical methods to specific clinical problems. 

\section{Conclusion} 

This study demonstrates that semi-supervised machine learning of in-home sensor data is effective at predicting ALSFRS-R scale trajectories, with incremental fine-tuned transfer learning performing well across functional domains. 
The findings indicate that selection of pseudo-labeling technique for estimating between-visit decline should be tailored to specific clinical objectives, with self-attention interpolation performing best for subscale-level monitoring and polynomial function interpolation for the summated composite ALSFRS-R score. 
However, the generalizability of reported modeled outcomes is constrained by limitations of the small participant cohort, reliance on bed sensor and motion detection data without comprehensive gait measurements that may be particularly important for assessing motor function. 
The low prediction error-low outcome correlation models for bulbar and motor-related subscale models (P1:Handwriting, Cutting, Stairs; P2: Swallowing; P3:Swallowing, Handwriting, Cutting, Stairs) exemplify the need for such measurements. 
Future research may validate the learning methods applied in this analysis with a larger, multi-center study to establish broader applicability, explore complementary clinical measures such as Forced Volume Capacity (FVC), and investigate enhanced feature engineering approaches that could improve performance for patient-heterogeneous ALSFRS-R component scales. 
Additionally, developing adaptive incremental learning algorithms with patient-specific clinical feedback mechanisms for ground truth scoring, and extending this framework through multi-model meta-learning approaches are promising directions for advancing personalized disease progression monitoring that could transform clinical decision making in neurodegenerative care. 

\bibliographystyle{abbrvnat}
\bibliography{citations}

\section*{Conflict of Interest Statement}
The authors declare that the research was conducted in the absence of any commercial or financial relationships that could be construed as a potential conflict of interest. 

\section*{Ethics Statement}
This work involved human subjects or animals in its research. 
Approval of all ethical and experimental procedures and protocols was granted by University of Missouri Institutional Review Board under Application No. 12345678. 

\section*{Author Contributions}
NM: Formal analysis, Investigation, Methodology, Validation, Writing – original draft, Writing – review \& editing. 
WJ: Conceptualization, Funding acquisition, Project administration, Supervision, Writing – review \& editing. 
MP: Conceptualization, Funding acquisition, Methodology, Writing – review \& editing. 
XS: Conceptualization, Funding acquisition, Methodology, Supervision, Writing – review \& editing. 

\section*{Funding}
This work was supported by the Department of Defense office of the Congressionally Directed
Medical Research Programs (CDMRP) through the Amyotrophic Lateral Sclerosis Research Program (ALSRP) Clinical Development Award (\#W81XWH-22-1-0491) and the ALS Association (\#24-AT-722). 
Opinions, interpretations, conclusions, and recommendations are those of the authors and are not necessarily endorsed by the Department of Defense. \\

\section*{Acknowledgments}
The University of Missouri has a financial relationship with Foresite Healthcare, who produces the sensor platform used in this study. 
Preparation of manuscript tables in LaTeX was assisted by Microsoft Copilot (version 1.25042.146) delivered via Microsoft 365 Copilot for Work. 
Thank you to Sheila Marushak and Zachary Selby. 

\section*{Supplemental Data}
Participant-level Taylor diagrams and supporting tabulated results may be found in the supplementary materials. 

\section*{Data Availability Statement}
The data supporting this study are derived from human subjects and contain sensitive information; therefore, they are not publicly available. 
Access to de-identified data may be granted upon reasonable request to [William Janes, Principal Investigator, janesw@health.missouri.edu], subject to institutional review board approval. 

\begin{table}[htbp]
\fontsize{10pt}{10pt}\selectfont
\renewcommand{\arraystretch}{1.1}
\caption{Participant enrollment and dataset characteristics.}
\label{tab:demographics_individual}
\centering
\setlength{\tabcolsep}{3pt}
\begin{tabular}{lccc}
    \toprule
     & \textbf{P1} & \textbf{P2} & \textbf{P3} \\
    \midrule
    \textbf{Age} (years)                 & 62           & 55           & 45 \\
    \textbf{Study Enrollment} (days)     & 599          & 236          & 219 \\
    \textbf{ALSFRS-R Instruments} (n)    & 15           & 8            & 8 \\
    \midrule
    \multicolumn{4}{l}{\textbf{Aggregate Dataset Lengths} (days)} \\
    \hspace{1em}Training                 & 389          & 128          & 156 \\
    \hspace{1em}Test                     & 98           & 33           & 40 \\
    \bottomrule
\end{tabular}
\end{table}

\begin{table}[h]
\centering
\fontsize{10pt}{10pt}\selectfont
\setlength{\tabcolsep}{3pt}
\begin{threeparttable}
  \caption{Hyperparameter search spaces for tuning XGBoost classifier algorithm.}
  \label{tab:hyperparameters}
  \begin{tabularx}{\columnwidth}{@{} l X l @{}}
    \toprule
    Hyperparameter & Description & Search Space \\
    \midrule
    eta (learning rate)      & Learning rate to scale the contribution of each tree.      
                             & [0.001, 0.01, 0.1, 0.3, 0.5] \\
    n\_estimators            & Number of boosting rounds (trees) to build.                
                             & [32, 64, 128, 192, 256, 384, 512]           \\
    gamma                    & Minimum loss reduction required to perform a split.        
                             & [0, 0.25, 0.5, 1] \\
    max\_depth               & Maximum depth of trees to prevent overfitting.             
                             & [2, 3, 4, 6, 8, 10, 12, 16, 24] \\
    min\_child\_weight       & Minimum sum of instance weights needed in a child node.    
                             & [0.5, 1, 3, 5, 7, 10] \\
    subsample                & Fraction of training data used for building each tree.     
                             & [0.8, 0.9, 1.0]   \\
    colsample\_bytree        & Fraction of features randomly sampled for each tree.       
                             & [0.6, 0.7, 0.8, 0.9] \\
    reg\_lambda              & L2 regularization to penalize large weights.               
                             & [0.01, 0.1, 1, 5, 10, 50, 100] \\
    reg\_alpha               & L1 regularization to encourage sparsity in feature weight. 
                             & [0, 0.001, 0.01, 0.1] \\
    \bottomrule
  \end{tabularx}
\end{threeparttable}
\end{table}

\begin{table}[htbp]
\centering
\fontsize{10pt}{10pt}\selectfont
\setlength{\tabcolsep}{1pt}
\caption{Observed variance in ALSFRS-R scales by participant.}
\label{tab:variance_scores}
\begin{tabular*}{\columnwidth}{@{\extracolsep{\fill}} l|ccc|ccc|ccc|ccc@{}}
  \toprule
    & \multicolumn{3}{c|}{\textbf{Bulbar}}
    & \multicolumn{3}{c|}{\textbf{Fine Motor}}
    & \multicolumn{3}{c|}{\textbf{Gross Motor}}
    & \multicolumn{3}{c}{\textbf{Respiratory}} \\
  \cmidrule(lr){2-4} \cmidrule(lr){5-7} \cmidrule(lr){8-10} \cmidrule(lr){11-13}
   \textbf{Ps}.\ & Speech & Saliv. & Swallow.
   & Handwrit. & Cut. & Dress.
   & Turn. & Walk. & Stairs
   & Dyspnea & Orthopnea & Resp. \\
  \midrule
  \textbf{P1} & 0.981 & 0     & 0.781 & 1.810 & 0.267 & 0.638 & 0.552 & 0.381 & 0.314 & 0.267 & 0.952 & 0 \\
  \textbf{P2} & 0.125 & 1.714 & 0.786 & 1.714 & 1.643 & 1.143 & 1.714 & 1.143 & 0.982 & 0.500 & 1.411 & 0.125 \\
  \textbf{P3} & 0.571 & 2.411 & 1.071 & 0.571 & 0.125 & 1.071 & 0.982 & 0.786 & 0.214 & 0.214 & 0.286 & 0.214 \\
  \bottomrule
\end{tabular*}
\end{table}

\begin{table}[htbp]
\fontsize{10pt}{10pt}\selectfont
\renewcommand{\arraystretch}{1.1}
\setlength{\tabcolsep}{2pt}
\caption{Mean model error (RMSE) and outcome correlation ($r$) across pseudo-labeling interpolation techniques by learning method and participant ALSFRS-R scale.}
\label{tab:result_learning_alsfrs}
\centering
\begin{tabular*}{\columnwidth}{@{\extracolsep{\fill}} 
 l c l | 
  S[table-format=1.2(1.2)] S[table-format=-1.2(1.2)] |
  S[table-format=1.2(1.2)] S[table-format=-1.2(1.2)] |
  S[table-format=1.2(1.2)] S[table-format=-1.2(1.2)]
}
\toprule
    &  & \textbf{Learning} 
    & \multicolumn{2}{c}{\textbf{Individual Batch}} 
    & \multicolumn{2}{c}{\textbf{Transfer Batch}} 
    & \multicolumn{2}{c}{\textbf{Transfer Incremental}} \\
\cmidrule(lr){3-3}\cmidrule(lr){4-5}\cmidrule(lr){6-7}\cmidrule(lr){8-9}
\textbf{Ps.} & Domain & Subscale 
    & \multicolumn{1}{c}{RMSE (SD)} & \multicolumn{1}{c}{$r$ (SD)} 
    & \multicolumn{1}{c}{RMSE (SD)} & \multicolumn{1}{c}{$r$ (SD)} 
    & \multicolumn{1}{c}{RMSE (SD)} & \multicolumn{1}{c}{$r$ (SD)} \\
\midrule
\textbf{P1} &             & Speech      & 0.20(0.07) & -0.05(0.09) & 0.08(0.02) & -0.02(0.03) & {\textbf{0.04}(0.01)} & 0.22(0.38) \\
   & Bulbar      & Salivation  & \multicolumn{1}{c}{--} & \multicolumn{1}{c}{--} & \multicolumn{1}{c}{--} & \multicolumn{1}{c}{--} & \multicolumn{1}{c}{--} & \multicolumn{1}{c}{--}       \\
   &             & Swallowing  & 0.34(0.15) & -0.05(0.21) & {\textbf{0.21}(0.09)} & 0.80(0.05)  & 0.23(0.06) & 0.62(0.41) \\
\cmidrule(lr){2-9}
   &             & Handwriting & 0.37(0.17) & -0.07(0.12) & 0.35(0.20) & -0.02(0.03) & {\textbf{0.15}(0.03)} & 0.08(0.14) \\
   & Fine Motor  & Cutting     & 0.14(0.04) & 0(0)  & {\textbf{0.09}(0.03)} & 0(0)  & 0.12(0.08) & 0.0(0.0) \\
   &             & Dressing    & 0.29(0.06) & 0.53(0.18)  & {\textbf{0.27}(0.05)} & 0.40(0.50)  & 0.35(0.07) & -0.20(0.13) \\
\cmidrule(lr){2-9}
   &             & Turning     & 0.27(0.05) & -0.07(0.35) & 0.32(0.03) & -0.03(0.08) & {\textbf{0.26}(0.02)} & 0.09(0.09) \\
   & Gross Motor & Walking     & {\textbf{0.05}(0.06)} & 0.10(0.18)  & 0.25(0.08) & 0.24(0.42)  & 0.26(0.10) & 0.24(0.42) \\
   &             & Stairs      & {\textbf{0.00}} & 0(0)  & 0.04(0) & 0(0)  & 0.08(0) & 0(0) \\
\cmidrule(lr){2-9}
   &             & Dyspnea     & \multicolumn{1}{c}{--} & \multicolumn{1}{c}{--} & 0.50(0.08) & 0.23(0.38)  & {\textbf{0.42}(0.10)} & 0.75(0.13) \\
   & Respiratory & Orthopnea   & 0.48(0.04) & 0.30(0.10) & 0.12(0.02) & 0.98(0.01)  & {\textbf{0.11}(0.01)} & 0.99(0) \\
   &             & Respiratory & \multicolumn{1}{c}{--} & \multicolumn{1}{c}{--} & \multicolumn{1}{c}{--} & \multicolumn{1}{c}{--} & \multicolumn{1}{c}{--} & \multicolumn{1}{c}{--}       \\
\cmidrule(lr){2-9}
   &             & Composite   & {\textbf{2.76}(1.70)} & 0.03(0.09)  & 3.29(1.05) & -0.04(0.68) & 3.24(0.19) & 0.57(0.12) \\
\midrule
\textbf{P2} &             & Speech      & \multicolumn{1}{c}{--} & \multicolumn{1}{c}{--} & 0.27(0.06) & 0.56(0.14)  & {\textbf{0.26}(0.05)} & 0.75(0.03) \\
   & Bulbar      & Salivation  & 0.49(0.04) & 0.09(0.25)  & 0.21(0.06) & -0.28(0.14) & {\textbf{0.20}(0.07)} & -0.20(0.13) \\
   &             & Swallowing  & 0.16(0.01) & 0(0)  & {\textbf{0.04}(0.01)} & 0(0)  & {\textbf{0.04}(0.01)} & 0(0)  \\
\cmidrule(lr){2-9}
   &             & Handwriting & 0.45(0.10) & 0.27(0.29)  & 0.12(0.04) & 0.32(0.31)  & {\textbf{0.10}(0.02)} & 0.46(0.35) \\
   & Fine Motor  & Cutting     & 0.37(0.07) & 0.15(0.34)  & 0.08(0.03) & -0.22(0.25) & {\textbf{0.05}(0.02)} & 0.28(0.34) \\
   &             & Dressing    & 0.35(0.04) & 0.21(0.23)  & 0.11(0.01) & 0.94(0.02)  & {\textbf{0.06}(0.02)} & 0.99(0.01) \\
\cmidrule(lr){2-9}
   &             & Turning     & 0.30(0) & 0.21(0.21)  & {\textbf{0.03}(0.02)} & 0.24(0.33)  & 0.12(0.03) & 0.67(0.21) \\
   & Gross Motor & Walking     & 0.30(0.04) & 0.20(0.13)  & 0.14(0.05) & 0.93(0.05)  & {\textbf{0.06}(0.01)} & 0.98(0.02) \\
   &             & Stairs      & 0.30(0.08) & 0.46(0.17)  & 0.24(0.01) & 0.88(0.08)  & {\textbf{0.23}(0.01)} & 0.92(0.03) \\
\cmidrule(lr){2-9}
   &             & Dyspnea     & \multicolumn{1}{c}{--} & \multicolumn{1}{c}{--} & {\textbf{0.46}(0.07)} & 0.88(0.03)  & 0.54(0.10) & 0.94(0.02) \\
   & Respiratory & Orthopnea   & 0.34(0.07) & 0.55(0.06)  & 0.31(0.02) & 0.67(0.16)  & {\textbf{0.30}(0.02)} & 0.73(0.12) \\
   &             & Respiratory & \multicolumn{1}{c}{--} & \multicolumn{1}{c}{--} & {\textbf{0.30}(0.11)} & 0.28(0.47)  & 0.31(0.11) & 0.48(0.17) \\
\cmidrule(lr){2-9}
   &             & Composite   & {\textbf{3.52}(0.77)} & 0.48(0.17)  & 6.03(1.95) & 0.11(0.20)  & 4.41(0.43) & 0.63(0.04) \\
\midrule
\textbf{P3} &             & Speech      & 0.38(0.15) & -0.13(0.10) & {\textbf{0.15}(0.03)} & 0(0.49)  & 0.19(0.02) & 0.04(0.39) \\
   & Bulbar      & Salivation  & 0.20(0.17) & -0.05(0.24) & {\textbf{0.15}(0.05)} & 0(0.01)  & 0.34(0.03) & 0.03(0.05) \\
   &             & Swallowing  & 0.12(0.12) & 0(0)  & {\textbf{0.10}(0.01)} & 0(0)  & 0.16(0.01) & 0(0) \\
\cmidrule(lr){2-9}
   &             & Handwriting & 0.11(0.10) & 0(0)  & {\textbf{0.06}(0.01)} & 0(0)  & 0.10(0.02) & 0(0) \\
   & Fine Motor  & Cutting     & {\textbf{0.02}} & 0(0)  & {\textbf{0.02}} & 0(0)  & 0.03(0) & 0(0) \\
   &             & Dressing    & 0.35(0.15) & -0.20(0.20) & 0.16(0.03) & 0.31(0.39)  & {\textbf{0.15}(0.02)} & 0.37(0.37) \\
\cmidrule(lr){2-9}
   &             & Turning     & {\textbf{0.15}(0.07)} & -0.08(0.14) & 0.16(0.02) & 0.08(0.14)  & 0.22(0.06) & 0.11(0.18) \\
   & Gross Motor & Walking     & 0.39(0.15) & -0.18(0.09) & 0.17(0.03) & -0.38(0.05) & {\textbf{0.16}(0.03)} & -0.37(0.06) \\
   &             & Stairs      & 0.09(0.08) & 0(0)  & {\textbf{0.03}(0.02)} & 0(0)  & 0.19(0.12) & 0(0) \\
\cmidrule(lr){2-9}
   &             & Dyspnea     & 0.19(0)  & -0.30(0)      & {\textbf{0.14}(0.06)} & -0.03(0.09) & 0.36(0.04) & 0.47(0.25) \\
   & Respiratory & Orthopnea   & {\textbf{0.10}(0.09)} & 0.39(0.34)  & 0.12(0.06) & 0.60(0.52)  & 0.22(0.11) & 0.64(0.55) \\
   &             & Respiratory & 0.22(0)  & -0.31(0)      & 0.16(0.06) & -0.27(0.06) & {\textbf{0.12}(0.04)} & 0.79(0.04) \\
\cmidrule(lr){2-9}
   &             & Composite   & 3.17(1.27) & -0.18(0.13) & 3.01(0.58) & -0.54(0.04) & {\textbf{2.97}(1.27)} & -0.41(0.43) \\
\bottomrule
\end{tabular*}
  \begin{tablenotes}
    \item \textbf{Bold} indicates lowest performing model error for ALSFRS-R subscale.
  \end{tablenotes}
\end{table}

\begin{table}[htbp]
\fontsize{10pt}{10pt}\selectfont
\renewcommand{\arraystretch}{1.1}
\setlength{\tabcolsep}{3pt}
\caption{Mean model error (RMSE) and outcome correlation ($r$) across participants and pseudo-labeling interpolation techniques by learning method and ALSFRS-R scale.}
\label{tab:result_learning_avg_alsfrs}
\centering
\begin{tabular*}{\columnwidth}{@{\extracolsep{\fill}} 
  c l |
  S[table-format=1.2(1.2)] S[table-format=-1.2(1.2)] |
  S[table-format=1.2(1.2)] S[table-format=-1.2(1.2)] |
  S[table-format=1.2(1.2)] S[table-format=-1.2(1.2)]
}
\toprule
  & \textbf{Learning}
  & \multicolumn{2}{c}{\textbf{Individual Batch}}
  & \multicolumn{2}{c}{\textbf{Transfer Batch}}
  & \multicolumn{2}{c}{\textbf{Transfer Incremental}} \\
\cmidrule(lr){2-2}\cmidrule(lr){3-4}\cmidrule(lr){5-6}\cmidrule(lr){7-8}
Domain & Subscale 
  & \multicolumn{1}{c}{RMSE (SD)} & \multicolumn{1}{c}{$r$ (SD)} 
  & \multicolumn{1}{c}{RMSE (SD)} & \multicolumn{1}{c}{$r$ (SD)} 
  & \multicolumn{1}{c}{RMSE (SD)} & \multicolumn{1}{c}{$r$ (SD)} \\
\midrule
            & Speech      & 0.29(0.11) & -0.09(0.10) & 0.17(0.04) &  0.18(0.22) & 0.16(0.03) &  0.34(0.27) \\
Bulbar      & Salivation  & 0.34(0.11) &  0.02(0.24) & 0.18(0.06) & -0.14(0.08) & 0.27(0.05) & -0.09(0.09) \\
            & Swallowing  & 0.21(0.09) & -0.02(0.07) & 0.12(0.04) &  0.27(0.02) & 0.14(0.03) &  0.21(0.14) \\
\cmidrule(lr){1-8}
            & Handwriting & 0.31(0.12) &  0.07(0.14) & 0.18(0.08) &  0.10(0.11) & 0.12(0.02) &  0.18(0.16) \\
Fine Motor  & Cutting     & 0.18(0.04) &  0.05(0.11) & 0.06(0.02) & -0.07(0.08) & 0.07(0.03) &  0.09(0.11) \\
            & Dressing    & 0.33(0.08) &  0.18(0.20) & 0.18(0.03) &  0.55(0.30) & 0.19(0.04) &  0.39(0.17) \\
\cmidrule(lr){1-8}
 & Turning  & 0.24(0.04) &  0.02(0.23) & 0.17(0.02) &  0.10(0.18) & 0.20(0.04) &  0.29(0.16) \\
Gross Motor & Walking     & 0.25(0.08) &  0.04(0.13) & 0.19(0.05) &  0.26(0.17) & 0.16(0.05) &  0.28(0.17) \\
            & Stairs      & 0.13(0.06) &  0.15(0.06) & 0.10(0.01) &  0.29(0.03) & 0.17(0.04) &  0.31(0.01) \\
\cmidrule(lr){1-8}
            & Dyspnea     & 0.19(0)    & -0.30(0)    & 0.37(0.07) &  0.36(0.17) & 0.44(0.08) &  0.72(0.13) \\
Respiratory & Orthopnea   & 0.31(0.07) &  0.41(0.17) & 0.18(0.03) &  0.75(0.23) & 0.22(0.05) &  0.79(0.22) \\
            & Respiratory & 0.22(0)    & -0.31(0)    & 0.23(0.08) &  0.01(0.27) & 0.21(0.07) &  0.64(0.11) \\
\cmidrule(lr){1-8}
            & Mean        & 0.25(0.07) & 0.02(0.12) & 0.18(0.04) & 0.22(0.15) & 0.20(0.04) & 0.35(0.15) \\
\cmidrule(lr){1-8}
       & Composite   & 3.15(1.25) &  0.11(0.13) & 4.11(1.19) & -0.16(0.31) & 3.54(0.63) &  0.26(0.20) \\
\bottomrule
\end{tabular*}
\end{table}

\begin{table}[htbp]
\fontsize{10pt}{10pt}\selectfont
\renewcommand{\arraystretch}{1.1}
\setlength{\tabcolsep}{2pt}
\caption{Mean model error (RMSE) and outcome correlation ($r$) across learning methods by pseudo-labeling interpolation technique and participant ALSFRS-R scale.}
\label{tab:result_interpolation_alsfrs}
\centering
\begin{tabular*}{\columnwidth}{@{\extracolsep{\fill}} 
  l c l |
  S[table-format=1.2(1.2)] S[table-format=-1.2(1.2)] |
  S[table-format=1.2(1.2)] S[table-format=-1.2(1.2)] |
  S[table-format=1.2(1.2)] S[table-format=-1.2(1.2)]
}
\toprule
  & & \textbf{Interpolation} 
  & \multicolumn{2}{c}{\textbf{Linear Slope}} 
  & \multicolumn{2}{c}{\textbf{Cubic Polynomial}} 
  & \multicolumn{2}{c}{\textbf{Self-Attention Ensemble}} \\
\cmidrule(lr){3-3}\cmidrule(lr){4-5}\cmidrule(lr){6-7}\cmidrule(lr){8-9}
\textbf{Ps.} & Domain & Subscale 
  & \multicolumn{1}{c}{RMSE (SD)} & \multicolumn{1}{c}{$r$ (SD)} 
  & \multicolumn{1}{c}{RMSE (SD)} & \multicolumn{1}{c}{$r$ (SD)} 
  & \multicolumn{1}{c}{RMSE (SD)} & \multicolumn{1}{c}{$r$ (SD)} \\
\midrule
\textbf{P1} &             & Speech      & 0.14(0.13) & 0(0)  & {\textbf{0.09}(0.05)}  & 0(0)  & 0.10(0.08)  & 0.15(0.44) \\
   & Bulbar      & Salivation  & \multicolumn{1}{c}{--} & \multicolumn{1}{c}{--} & \multicolumn{1}{c}{--} & \multicolumn{1}{c}{--} & \multicolumn{1}{c}{--} & \multicolumn{1}{c}{--} \\
   &             & Swallowing  & 0.27(0.05) & 0.49(0.62)  & 0.35(0.12)  & 0.52(0.55)  & {\textbf{0.16}(0.04)}  & 0.36(0.34) \\
\cmidrule(lr){2-9}
   &             & Handwriting & 0.39(0.19) & 0(0)  & 0.33(0.15)  & 0(0)  & {\textbf{0.14}(0.05)}  & 0(0.23) \\
   & Fine Motor  & Cutting     & 0.15(0.03) & 0(0)  & 0.12(0.05)  & 0(0)  & {\textbf{0.07}(0.04)}  & 0(0) \\
   &             & Dressing    & 0.32(0.03) & 0.19(0.40)  & 0.32(0.11)  & 0.46(0.47)  & {\textbf{0.27}(0.05)}  & 0.09(0.52) \\
\cmidrule(lr){2-9}
   &             & Turning     & 0.31(0.04) & -0.01(0.15) & {\textbf{0.26}(0.05)}  & -0.10(0.27) & 0.28(0.01)  & 0.11(0.15) \\
   & Gross Motor & Walking     & 0.21(0.17) & 0(0)  & 0.21(0.17)  & 0(0)  & {\textbf{0.14}(0.02)}  & 0.59(0.24) \\
   &             & Stairs      & 0.04(0.04) & 0(0)  & 0.04(0.04)  & 0(0)  & \multicolumn{1}{c}{--} & \multicolumn{1}{c}{--} \\
\cmidrule(lr){2-9}
   &             & Dyspnea     & 0.52(0.04) & 0.68(0.24)  & 0.50(0.05)  & 0.59(0.28)  & {\textbf{0.36}(0.08)}  & 0.20(0.57) \\
   & Respiratory & Orthopnea   & 0.26(0.23) & 0.79(0.34)  & {\textbf{0.22}(0.21)}  & 0.76(0.39)  & 0.23(0.19)  & 0.73(0.46) \\
   &             & Respiratory & \multicolumn{1}{c}{--} & \multicolumn{1}{c}{--} & \multicolumn{1}{c}{--} & \multicolumn{1}{c}{--} & \multicolumn{1}{c}{--} & \multicolumn{1}{c}{--} \\
\cmidrule(lr){2-9}
   &             & Composite   & {\textbf{2.79}(0.84)} & 0.08(0.44)  & 2.38(0.83)  & -0.02(0.50) & 4.11(0.67)  & 0.50(0.38) \\
\midrule
\textbf{P2} &             & Speech      & 0.30(0.01) & 0.59(0.18)  & 0.30(0.01) & 0.62(0.18)  & {\textbf{0.20}} & 0.75(0.04) \\
   & Bulbar      & Salivation  & 0.29(0.20) & -0.13(0.18) & {\textbf{0.28}(0.20)} & -0.29(0.14) & 0.33(0.10) & 0.04(0.27) \\
   &             & Swallowing  & 0.09(0.07) & 0(0)  & {\textbf{0.08}(0.08)} & 0(0)  & {\textbf{0.08}(0.06)} & 0(0) \\
\cmidrule(lr){2-9}
   &             & Handwriting & 0.23(0.25) & 0.24(0.22)  & 0.25(0.22) & 0.13(0.13)  & {\textbf{0.20}(0.13)} & 0.67(0.15) \\
   & Fine Motor  & Cutting     & 0.15(0.19) & -0.01(0.27) & 0.20(0.20) & 0(0.06)  & {\textbf{0.14}(0.14)} & 0.21(0.60) \\
   &             & Dressing    & {\textbf{0.16}(0.14)} & 0.80(0.30)  & 0.17(0.18) & 0.69(0.49)  & 0.19(0.15) & 0.64(0.53) \\
\cmidrule(lr){2-9}
   &             & Turning     & 0.15(0.14) & 0.36(0.23)  & 0.16(0.13) & 0.35(0.24)  & {\textbf{0.13}(0.15)} & 0.41(0.53) \\
   & Gross Motor & Walking     & 0.17(0.13) & 0.75(0.38)  & 0.20(0.13) & 0.70(0.41)  & {\textbf{0.14}(0.10)} & 0.66(0.53) \\
   &             & Stairs      & 0.28(0.07) & 0.73(0.34)  & 0.27(0.05) & 0.76(0.31)  & {\textbf{0.22}(0.01)} & 0.78(0.12) \\
\cmidrule(lr){2-9}
   &             & Dyspnea     & 0.55(0.11) & 0.92(0.04)  & 0.52(0.02) & 0.90(0.08)  & {\textbf{0.41}(0.04)} & 0.91(0.01) \\
   & Respiratory & Orthopnea   & {\textbf{0.29}(0.01)} & 0.62(0.06)  & 0.31(0) & 0.58(0.08)  & 0.36(0.05) & 0.75(0.19) \\
   &             & Respiratory & 0.34(0.01) & 0.62(0.02)  & 0.40(0) & 0.48(0)  & {\textbf{0.19}} & 0.02(0.40) \\
\cmidrule(lr){2-9}
   &             & Composite   & {\textbf{4.21}(1.05)} & 0.39(0.20)  & 4.22(0.94) & 0.42(0.19)  & 5.54(2.39) & 0.41(0.46) \\
\midrule
\textbf{P3} &             & Speech      & 0.25(0.17) & 0.13(0.22)  & 0.27(0.20)  & 0.11(0.30)  & {\textbf{0.20}(0.01)}  & -0.34(0.27) \\
   & Bulbar      & Salivation  & 0.20(0.14) & -0.07(0.13) & {\textbf{0.19}(0.15)}  & -0.05(0.09) & 0.30(0.10)  & 0.10(0.12)  \\
   &             & Swallowing  & 0.12(0.03) & 0(0)  & 0.16(0.07)  & 0(0)  & {\textbf{0.09}(0.09)}  & 0(0)  \\
\cmidrule(lr){2-9}
   &             & Handwriting & 0.11(0.06) & 0(0)  & 0.12(0.06)  & 0(0)  & {\textbf{0.05}(0.04)}  & 0(0)  \\
   & Fine Motor  & Cutting     & {\textbf{0.02}(0.01)} & 0(0)  & {\textbf{0.02}(0.01)}  & 0(0)  & \multicolumn{1}{c}{--} & \multicolumn{1}{c}{--} \\
   &             & Dressing    & 0.25(0.14) & -0.06(0.20) & 0.27(0.17)  & 0.01(0.32)  & 0.14(0.03)  & 0.52(0.43)  \\
\cmidrule(lr){2-9}
   &             & Turning     & {\textbf{0.17}(0.02)} & 0(0)  & 0.18(0.03)  & 0(0)  & 0.18(0.11)  & 0.10(0.31)  \\
   & Gross Motor & Walking     & 0.24(0.16) & -0.29(0.06) & 0.27(0.21)  & -0.27(0.17) & {\textbf{0.21}(0.01)}  & -0.37(0.10) \\
   &             & Stairs      & 0.13(0.12) & 0(0)  & 0.15(0.12)  & 0(0)  & {\textbf{0.04}(0.03)}  & 0(0)  \\
\cmidrule(lr){2-9}
   &             & Dyspnea     & {\textbf{0.24}(0.20)} & 0.35(0.40)  & 0.25(0.20)  & 0.26(0.47)  & {\textbf{0.24}(0.07)}  & -0.07(0.24) \\
   & Respiratory & Orthopnea   & 0.20(0.11) & 0.82(0.21)  & 0.19(0.04)  & 0.81(0.19)  & {\textbf{0.05}(0.06)}  & 0(0)  \\
   &             & Respiratory & {\textbf{0.10}(0.01)} & 0.24(0.74)  & 0.12(0.03)  & 0.29(0.69)  & 0.20(0.03)  & 0.07(0.66)  \\
\cmidrule(lr){2-9}
   &             & Composite   & {\textbf{2.38}(0.19)} & -0.52(0.21) & 2.53(0.29)  & -0.46(0.23) & 4.24(0.50)  & -0.15(0.31) \\
\bottomrule
\end{tabular*}
  \begin{tablenotes}
    \item \textbf{Bold} indicates lowest performing model error for ALSFRS-R subscale.
  \end{tablenotes}
\end{table}

\begin{table}[htbp]
\fontsize{10pt}{10pt}\selectfont
\renewcommand{\arraystretch}{1.1}
\setlength{\tabcolsep}{3pt}
\caption{Mean model error (RMSE) and outcome correlation ($r$) across participants and learning methods by pseudo-labeling interpolation technique and ALSFRS-R scale.}
\label{tab:result_interpolation_avg_alsfrs}
\centering
\begin{tabular*}{\columnwidth}{@{\extracolsep{\fill}} 
  c l |
  S[table-format=1.2(1.2)] S[table-format=1.2(1.2)] |
  S[table-format=1.2(1.2)] S[table-format=1.2(1.2)] |
  S[table-format=1.2(1.2)] S[table-format=1.2(1.2)]
}
\toprule
  & \textbf{Interpolation }
  & \multicolumn{2}{c}{\textbf{Linear Slope}} 
  & \multicolumn{2}{c}{\textbf{Cubic Polynomial}} 
  & \multicolumn{2}{c}{\textbf{Self-Attention Ensemble}} \\
\cmidrule(lr){2-2}\cmidrule(lr){3-4}\cmidrule(lr){5-6}\cmidrule(lr){7-8}
Domain & Subscale 
  & \multicolumn{1}{c}{RMSE (SD)} & \multicolumn{1}{c}{$r$ (SD)} 
  & \multicolumn{1}{c}{RMSE (SD)} & \multicolumn{1}{c}{$r$ (SD)} 
  & \multicolumn{1}{c}{RMSE (SD)} & \multicolumn{1}{c}{$r$ (SD)} \\
\midrule
            & Speech      & 0.23(0.10) &  0.24(0.13) & 0.22(0.09) &  0.24(0.16) & 0.17(0.03) &  0.19(0.25) \\
Bulbar      & Salivation  & 0.24(0.17) & -0.10(0.15) & 0.24(0.17) & -0.17(0.12) & 0.32(0.10) &  0.07(0.20) \\
            & Swallowing  & 0.16(0.05) &  0.16(0.21) & 0.20(0.09) &  0.17(0.18) & 0.11(0.06) &  0.12(0.11) \\
\cmidrule(lr){1-8}
            & Handwriting & 0.24(0.17) &  0.08(0.07) & 0.23(0.14) &  0.04(0.04) & 0.13(0.07) &  0.22(0.13) \\
Fine Motor  & Cutting     & 0.11(0.08) & -0.00(0.09) & 0.11(0.09) &  0.00(0.02) & 0.11(0.09) &  0.10(0.30) \\
            & Dressing    & 0.24(0.10) &  0.31(0.30) & 0.25(0.15) &  0.39(0.43) & 0.20(0.08) &  0.42(0.49) \\
\cmidrule(lr){1-8}
            & Turning     & 0.21(0.07) &  0.12(0.13) & 0.20(0.07) &  0.08(0.17) & 0.20(0.09) &  0.21(0.33) \\
Gross Motor & Walking     & 0.21(0.15) &  0.15(0.15) & 0.23(0.17) &  0.14(0.19) & 0.16(0.04) &  0.29(0.29) \\
            & Stairs      & 0.15(0.08) &  0.24(0.11) & 0.15(0.07) &  0.25(0.10) & 0.13(0.02) &  0.39(0.06) \\
\cmidrule(lr){1-8}
            & Dyspnea     & 0.44(0.12) &  0.65(0.23) & 0.42(0.09) &  0.58(0.28) & 0.34(0.06) &  0.35(0.27) \\
Respiratory & Orthopnea   & 0.25(0.12) &  0.74(0.20) & 0.24(0.08) &  0.72(0.22) & 0.21(0.10) &  0.49(0.22) \\
            & Respiratory & 0.22(0.01) &  0.43(0.38) & 0.26(0.01) &  0.39(0.34) & 0.20(0.01) &  0.05(0.53) \\
\cmidrule(lr){1-8}
            & Mean        & 0.23(0.10) & 0.25(0.18) & 0.23(0.10) & 0.24(0.19) & 0.19(0.06) & 0.24(0.27) \\
\cmidrule(lr){1-8}
            & Composite   & 3.13(0.69) & -0.02(0.28) & 3.04(0.69) & -0.02(0.31) & 4.63(1.19) &  0.25(0.38) \\
\bottomrule
\end{tabular*}
\end{table}

\end{document}



\title{Enhancing ALS Progression Tracking with Semi-Supervised ALSFRS-R Scores Estimated from Ambient Home Health Monitoring}
\maketitle

\section{Supplementary Figures}

\begin{figure}[h!]
\begin{center}
\includegraphics[width=\textwidth]{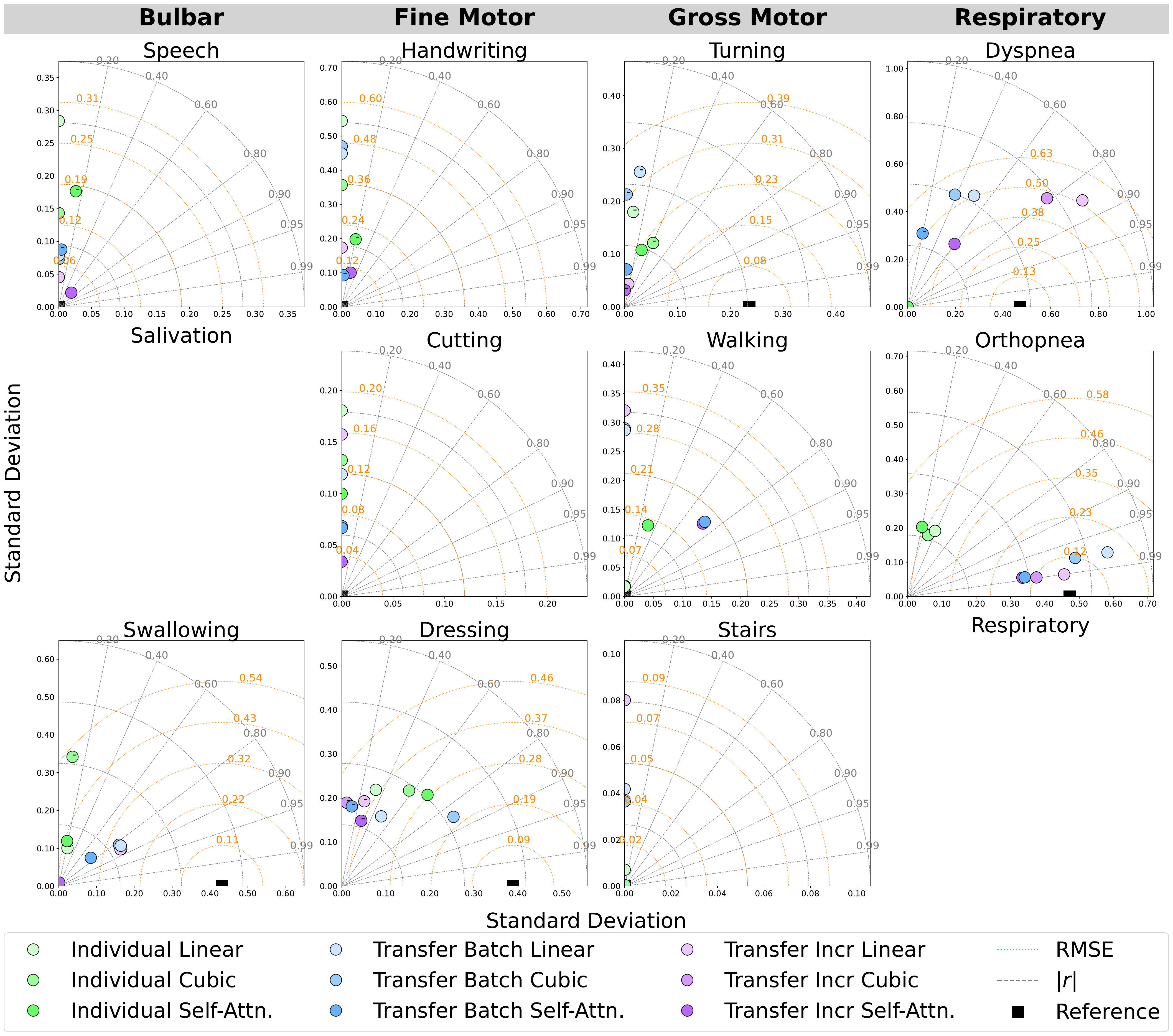}
\end{center}
\caption{Participant 1 Taylor diagrams showing mean prediction error (RMSE), absolute correlation ($|r|$), and standard deviation of outcomes for each ALSFRS-R scale, annotated by negative ($-$) correlation.}
\label{fig:TD_each_userid_nonorm_1}
\end{figure}

\begin{figure}[h!]
\begin{center}
\includegraphics[width=\textwidth]{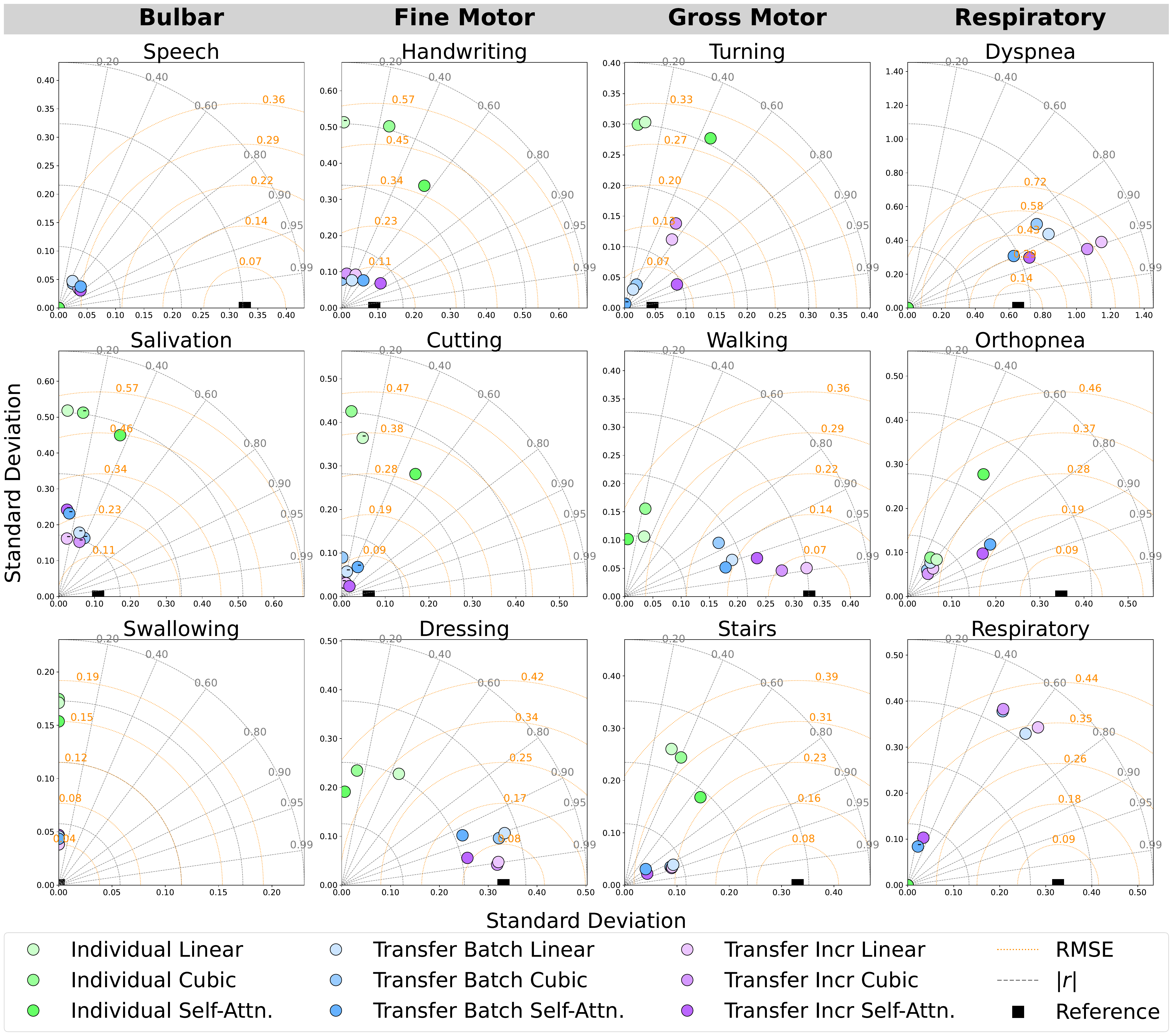}
\end{center}
\caption{Participant 2 Taylor diagrams showing mean prediction error (RMSE), absolute correlation ($|r|$), and standard deviation of outcomes for each ALSFRS-R scale, annotated by negative ($-$) correlation.}
\label{fig:TD_each_userid_nonorm_2}
\end{figure}

\begin{figure}[h!]
\begin{center}
\includegraphics[width=\textwidth]{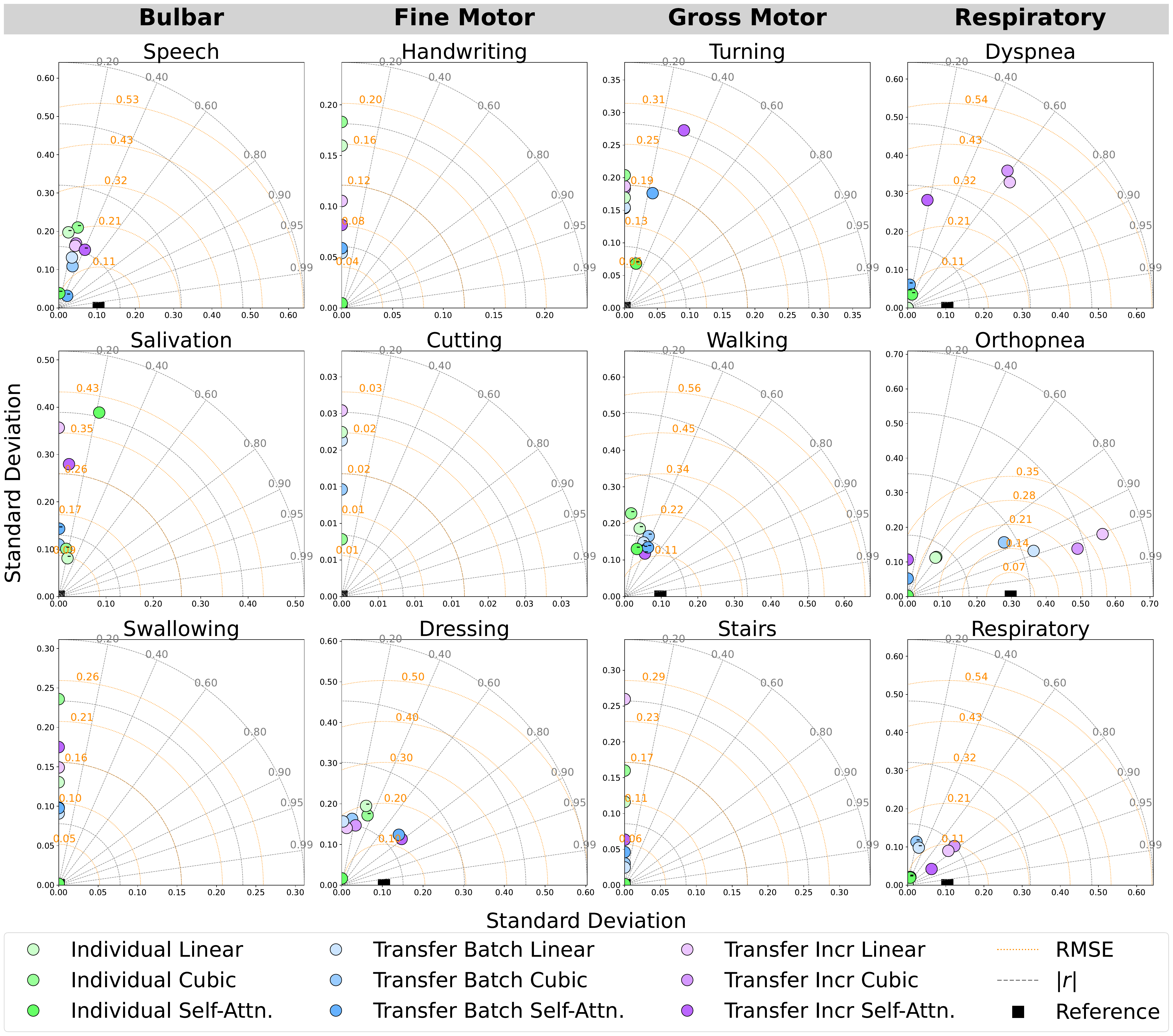}
\end{center}
\caption{Participant 3 Taylor diagrams showing mean prediction error (RMSE), absolute correlation ($|r|$), and standard deviation of outcomes for each ALSFRS-R scale, annotated by negative ($-$) correlation.}
\label{fig:TD_each_userid_nonorm_3}
\end{figure}

\clearpage

\section{Supplementary Tables}

\begin{table}[h!]
\fontsize{10pt}{10pt}\selectfont
\setlength{\tabcolsep}{2pt}
\caption{Individual batch mean model error (RMSE) and outcome correlation ($r$) by pseudo-labeling interpolation technique and participant ALSFRS-R scale.}
\label{tab:result_individual_batch_alsfrs}
\centering
\begin{tabular*}{\columnwidth}{@{\extracolsep{\fill}} 
  l c l |
  S[table-format=1.2] S[table-format=1.2] |
  S[table-format=1.2] S[table-format=1.2] |
  S[table-format=1.2] S[table-format=1.2]
}
\toprule
  & & \textbf{Interpolation} 
  & \multicolumn{2}{c}{\centering \textbf{Linear Slope}} 
  & \multicolumn{2}{c}{\centering \textbf{Cubic Polynomial}} 
  & \multicolumn{2}{c}{\centering \textbf{Self-Attention Ensemble}} \\
\cmidrule(lr){3-3}\cmidrule(lr){4-5}\cmidrule(lr){6-7}\cmidrule(lr){8-9}
\textbf{Ps.} & Domain & Subscale 
    & \multicolumn{1}{c}{RMSE} & \multicolumn{1}{c}{$r$} 
    & \multicolumn{1}{c}{RMSE} & \multicolumn{1}{c}{$r$} 
    & \multicolumn{1}{c}{RMSE} & \multicolumn{1}{c}{$r$} \\
\midrule
\textbf{P1}  &             & Speech      & 0.28 & 0.00  & 0.14 & 0.00  & 0.18 & -0.15 \\
    & Bulbar      & Salivation  & {--} & {--}  & {--} & {--}  & {--} & {--} \\
    &             & Swallowing  & 0.33 & -0.22 & 0.49 & -0.11 & 0.19 & 0.18 \\
\cmidrule(lr){2-9}
    &             & Handwriting & 0.54 & 0.00  & 0.36 & 0.00  & 0.20 & -0.20 \\
    & Fine Motor  & Cutting     & 0.18 & 0.00  & 0.13 & 0.00  & 0.10 & 0.00 \\
    &             & Dressing    & 0.33 & 0.33  & 0.32 & 0.58  & 0.22 & 0.68 \\
\cmidrule(lr){2-9}
    &             & Turning     & 0.32 & -0.09 & 0.23 & -0.41 & 0.27 & 0.28 \\
    & Gross Motor & Walking     & 0.02 & 0.00  & 0.02 & 0.00  & 0.12 & 0.31 \\
    &             & Stairs      & 0.01 & 0.00  & 0.00 & 0.00  & {--} & {--} \\
\cmidrule(lr){2-9}
    &             & Dyspnea     & {--} & {--}  & {--} & {--}  & {--} & {--} \\
    & Respiratory & Orthopnea   & 0.53 & 0.39  & 0.47 & 0.31  & 0.45 & 0.20 \\
    &             & Respiratory & {--} & {--}  & {--} & {--}  & {--} & {--} \\    
\cmidrule(lr){2-9}
    & Composite   & Composite   & 1.86 & 0.10  & 1.69 & -0.07 & 4.72 & 0.06 \\
\midrule
\textbf{P2}  &             & Speech      & {--} & {--}  & {--} & {--}  & {--} & {--}  \\
    & Bulbar      & Salivation  & 0.52 & 0.05  & 0.51 & -0.13 & 0.45 & 0.36 \\
    &             & Swallowing  & 0.17 & 0.00  & 0.17 & 0.00  & 0.15 & 0.00 \\
\cmidrule(lr){2-9}
    &             & Handwriting & 0.51 & -0.01 & 0.50 & 0.25 & 0.34 & 0.56 \\
    & Fine Motor  & Cutting     & 0.37 & -0.13 & 0.43 & 0.05 & 0.30 & 0.52 \\
    &             & Dressing    & 0.31 & 0.46  & 0.38 & 0.13 & 0.36 & 0.03 \\
\cmidrule(lr){2-9}
    &             & Turning     & 0.30 & 0.11  & 0.30 & 0.07 & 0.30 & 0.45 \\
    & Gross Motor & Walking     & 0.31 & 0.31  & 0.33 & 0.23 & 0.26 & 0.05 \\
    &             & Stairs      & 0.36 & 0.33  & 0.33 & 0.40 & 0.21 & 0.65 \\
\cmidrule(lr){2-9}
    &  & Dyspnea  & {--} & {--}  & {--} & {--}  & {--} & {--} \\
    & Respiratory & Orthopnea   & 0.29 & 0.62  & 0.31 & 0.50 & 0.42 & 0.53 \\
    &             & Respiratory & {--} & {--}  & {--} & {--}  & {--} & {--} \\  
\cmidrule(lr){2-9}
    & Composite   & Composite   & 3.01 & 0.34 & 3.14 & 0.43 & 4.41 & 0.66 \\
\midrule
\textbf{P3}  &             & Speech      & 0.45 & -0.13 & 0.49 & -0.23 & 0.21 & -0.04 \\
    & Bulbar      & Salivation  & 0.10 & -0.22 & 0.10 & -0.16 & 0.39 & 0.22 \\
    &             & Swallowing  & 0.13 & 0.00  & 0.24 & 0.00  & 0.00 & 0.00 \\
\cmidrule(lr){2-9}
    &             & Handwriting & 0.16 & 0.00  & 0.18 & 0.00  & 0.00 & 0.00 \\
    & Fine Motor  & Cutting     & 0.02 & 0.00  & 0.01 & 0.00  & {--} & {--} \\
    &             & Dressing    & 0.41 & -0.29 & 0.46 & -0.35 & 0.18 & 0.03 \\
\cmidrule(lr){2-9}
    &             & Turning     & 0.17 & 0.00  & 0.20 & 0.00  & 0.07 & -0.25 \\
    & Gross Motor & Walking     & 0.43 & -0.22 & 0.51 & -0.08 & 0.22 & -0.25 \\
    &             & Stairs      & 0.12 & 0.00  & 0.16 & 0.00  & 0.00 & 0.00 \\
\cmidrule(lr){2-9}
    &             & Dyspnea     & {--} & {--}  & {--} & {--}  & 0.19 & -0.30 \\
    & Respiratory & Orthopnea   & 0.13 & 0.58  & 0.18 & 0.59 & 0.00 & 0.00 \\
    &             & Respiratory & {--} & {--}  & {--} & {--}  & 0.22 & -0.31 \\
\cmidrule(lr){2-9}
    & Composite   & Composite   & 2.27 & -0.29 & 2.61 & -0.20 & 4.62 & -0.04 \\
\bottomrule
\end{tabular*}
\end{table}

\begin{table}[h!]
\fontsize{10pt}{10pt}\selectfont
\renewcommand{\arraystretch}{1}
\setlength{\tabcolsep}{2pt}
\caption{Transfer batch mean model error (RMSE) and outcome correlation ($r$) by pseudo-labeling interpolation technique and participant ALSFRS-R scale.}
\label{tab:result_transfer_batch_alsfrs}
\centering
\begin{tabular*}{\columnwidth}{@{\extracolsep{\fill}} 
  l c l |
  S[table-format=1.2] S[table-format=1.2] |
  S[table-format=1.2] S[table-format=1.2] |
  S[table-format=1.2] S[table-format=1.2]
}
\toprule
  & & \textbf{Interpolation} 
  & \multicolumn{2}{c}{\centering \textbf{Linear Slope}} 
  & \multicolumn{2}{c}{\centering \textbf{Cubic Polynomial}} 
  & \multicolumn{2}{c}{\centering \textbf{Self-Attention Ensemble}} \\
\cmidrule(lr){3-3}\cmidrule(lr){4-5}\cmidrule(lr){6-7}\cmidrule(lr){8-9}
\textbf{Ps.} & Domain & Subscale 
    & \multicolumn{1}{c}{RMSE (SD)} & \multicolumn{1}{c}{$r$ (SD)} 
    & \multicolumn{1}{c}{RMSE (SD)} & \multicolumn{1}{c}{$r$ (SD)} 
    & \multicolumn{1}{c}{RMSE (SD)} & \multicolumn{1}{c}{$r$ (SD)} \\
\midrule
\textbf{P1}  &             & Speech      & 0.08 & 0.00  & 0.07 & 0.00  & 0.10 & -0.05 \\
    & Bulbar      & Salivation  & {--} & {--}  & {--} & {--}  & {--} & {--} \\
    &             & Swallowing  & 0.24 & 0.84  & 0.29 & 0.82  & 0.11 & 0.75 \\
\cmidrule(lr){2-9}
    &             & Handwriting & 0.45 & 0.00  & 0.47 & 0.00  & 0.12 & -0.05 \\
    & Fine Motor  & Cutting     & 0.12 & 0.00  & 0.07 & 0.00  & 0.07 & 0.00 \\
    &             & Dressing    & 0.29 & 0.49  & 0.21 & 0.85  & 0.31 & -0.13 \\
\cmidrule(lr){2-9}
    &             & Turning     & 0.35 & -0.11 & 0.32 & -0.02 & 0.29 & 0.05 \\
    & Gross Motor & Walking     & 0.29 & 0.00  & 0.29 & 0.00  & 0.16 & 0.73 \\
    &             & Stairs      & 0.04 & 0.00  & 0.04 & 0.00  & {--} & {--} \\
\cmidrule(lr){2-9}
    &             & Dyspnea     & 0.54 & 0.51  & 0.54 & 0.39  & 0.41 & -0.20 \\
    & Respiratory & Orthopnea   & 0.14 & 0.98  & 0.11 & 0.97  & 0.12 & 0.99 \\
    &             & Respiratory & {--} & {--}  & {--} & {--}  & {--} & {--} \\    
\cmidrule(lr){2-9}
    & Composite   & Composite   & 3.50 & -0.36 & 2.15 & -0.49 & 4.21 & 0.74 \\
\midrule
\textbf{P2}  &             & Speech      & 0.30 & 0.46  & 0.31 & 0.50  & 0.20 & 0.72 \\
    & Bulbar      & Salivation  & 0.18 & -0.31 & 0.17 & -0.40 & 0.27 & -0.13 \\
    &             & Swallowing  & 0.05 & 0.00  & 0.04 & 0.00  & 0.04 & 0.00 \\
\cmidrule(lr){2-9}
    &             & Handwriting & 0.08 & 0.35  & 0.12 & 0.00  & 0.15 & 0.62 \\
    & Fine Motor  & Cutting     & 0.06 & -0.20 & 0.11 & 0.01  & 0.07 & -0.48 \\
    &             & Dressing    & 0.11 & 0.95  & 0.10 & 0.96  & 0.12 & 0.92 \\
\cmidrule(lr){2-9}
    &             & Turning     & 0.03 & 0.41  & 0.05 & 0.45  & 0.02 & -0.14 \\
    & Gross Motor & Walking     & 0.15 & 0.95  & 0.19 & 0.87  & 0.09 & 0.96 \\
    &             & Stairs      & 0.24 & 0.92  & 0.25 & 0.93  & 0.23 & 0.79 \\
\cmidrule(lr){2-9}
    &             & Dyspnea     & 0.48 & 0.89  & 0.51 & 0.84  & 0.38 & 0.90 \\
    & Respiratory & Orthopnea   & 0.30 & 0.56  & 0.31 & 0.59  & 0.33 & 0.85 \\
    &             & Respiratory & 0.34 & 0.61  & 0.40 & 0.48  & 0.19 & -0.26 \\    
\cmidrule(lr){2-9}
    & Composite   & Composite   & 4.97 & 0.21  & 4.85 & 0.23  & 8.28 & -0.12 \\
\midrule
\textbf{P3}  &             & Speech      & 0.14 & 0.25  & 0.13 & 0.31  & 0.19 & -0.57 \\
    & Bulbar      & Salivation  & 0.14 & 0.00  & 0.11 & 0.00  & 0.20 & -0.01 \\
    &             & Swallowing  & 0.09 & 0.00  & 0.10 & 0.00  & 0.10 & 0.00 \\
\cmidrule(lr){2-9}
    &             & Handwriting & 0.05 & 0.00  & 0.06 & 0.00  & 0.06 & 0.00 \\
    & Fine Motor  & Cutting     & 0.02 & 0.00  & 0.01 & 0.00  & {--} & {--} \\
    &             & Dressing    & 0.18 & 0.02  & 0.18 & 0.16  & 0.13 & 0.75 \\
\cmidrule(lr){2-9}
    &             & Turning     & 0.15 & 0.00  & 0.15 & 0.00  & 0.18 & 0.24 \\
    & Gross Motor & Walking     & 0.15 & -0.33 & 0.17 & -0.37 & 0.20 & -0.43 \\
    &             & Stairs      & 0.02 & 0.00  & 0.03 & 0.00  & 0.05 & 0.00 \\
\cmidrule(lr){2-9}
    &             & Dyspnea     & 0.10 & 0.07  & 0.11 & -0.08 & 0.21 & -0.09 \\
    & Respiratory & Orthopnea   & 0.14 & 0.94  & 0.16 & 0.87  & 0.05 & 0.00 \\
    &             & Respiratory & 0.11 & -0.28 & 0.14 & -0.20 & 0.22 & -0.32 \\    
\cmidrule(lr){2-9}
    & Composite   & Composite   & 2.60 & -0.58 & 2.77 & -0.54 & 3.67 & -0.50 \\
\bottomrule
\end{tabular*}
\end{table}

\begin{table}[h!]
\fontsize{10pt}{10pt}\selectfont
\renewcommand{\arraystretch}{1}
\setlength{\tabcolsep}{2pt}
\caption{Transfer incremental mean model error (RMSE) and outcome correlation ($r$) by pseudo-labeling interpolation technique and participant ALSFRS-R scale.}
\label{tab:result_transfer_incremental_alsfrs}
\centering
\begin{tabular*}{\columnwidth}{@{\extracolsep{\fill}} 
  l c l |
  S[table-format=1.2] S[table-format=1.2] |
  S[table-format=1.2] S[table-format=1.2] |
  S[table-format=1.2] S[table-format=1.2]
}
& & \textbf{Interpolation} 
& \multicolumn{2}{c}{\centering \textbf{Linear Slope}} 
& \multicolumn{2}{c}{\centering \textbf{Cubic Polynomial}} 
& \multicolumn{2}{c}{\centering \textbf{Self-Attention Ensemble}} \\
\cmidrule(lr){3-3}\cmidrule(lr){4-5}\cmidrule(lr){6-7}\cmidrule(lr){8-9}
\textbf{Ps.} & Domain & Subscale 
    & \multicolumn{1}{c}{RMSE (SD)} & \multicolumn{1}{c}{$r$ (SD)} 
    & \multicolumn{1}{c}{RMSE (SD)} & \multicolumn{1}{c}{$r$ (SD)} 
    & \multicolumn{1}{c}{RMSE (SD)} & \multicolumn{1}{c}{$r$ (SD)} \\
\midrule
\textbf{P1}  &             & Speech      & 0.05 & 0.00  & 0.05 & 0.00  & 0.03 & 0.66 \\
    & Bulbar      & Salivation  & {--} & {--}  & {--} & {--}  & {--} & {--} \\
    &             & Swallowing  & 0.24 & 0.86  & 0.28 & 0.86  & 0.17 & 0.15 \\
\cmidrule(lr){2-9}
    &             & Handwriting & 0.17 & 0.00  & 0.17 & 0.00  & 0.11 & 0.25 \\
    & Fine Motor  & Cutting     & 0.16 & 0.00  & 0.16 & 0.00  & 0.03 & 0.00 \\
    &             & Dressing    & 0.34 & -0.26 & 0.42 & -0.06 & 0.28 & -0.29 \\
\cmidrule(lr){2-9}
    &             & Turning     & 0.27 & 0.16  & 0.24 & 0.12  & 0.28 & -0.01 \\
    & Gross Motor & Walking     & 0.32 & 0.00  & 0.32 & 0.00  & 0.15 & 0.73 \\
    &             & Stairs      & 0.08 & 0.00  & 0.08 & 0.00  & {--} & {--} \\
\cmidrule(lr){2-9}
    &             & Dyspnea     & 0.49 & 0.85  & 0.47 & 0.79  & 0.30 & 0.60 \\
    & Respiratory & Orthopnea   & 0.11 & 0.99  & 0.11 & 0.99  & 0.13 & 0.99 \\
    &             & Respiratory & {--} & {--}  & {--} & {--}  & {--} & {--} \\    
\cmidrule(lr){2-9}
    & Composite   & Composite   & 3.02 & 0.51  & 3.31 & 0.50  & 3.39 & 0.71 \\
\midrule
\textbf{P2}  &             & Speech      & 0.29 & 0.72  & 0.29 & 0.75  & 0.20 & 0.78 \\
    & Bulbar      & Salivation  & 0.16 & -0.14 & 0.16 & -0.35 & 0.28 & -0.10 \\
    &             & Swallowing  & 0.04 & 0.00  & 0.04 & 0.00  & 0.05 & 0.00 \\
\cmidrule(lr){2-9}
    &             & Handwriting & 0.09 & 0.39  & 0.12 & 0.14  & 0.10 & 0.84 \\
    & Fine Motor  & Cutting     & 0.03 & 0.30  & 0.07 & -0.07 & 0.05 & 0.60 \\
    &             & Dressing    & 0.05 & 0.99  & 0.04 & 0.99  & 0.08 & 0.98 \\
\cmidrule(lr){2-9}
    &             & Turning     & 0.13 & 0.57  & 0.14 & 0.52  & 0.08 & 0.91 \\
    & Gross Motor & Walking     & 0.05 & 0.99  & 0.07 & 0.99  & 0.07 & 0.96 \\
    &             & Stairs      & 0.24 & 0.93  & 0.24 & 0.94  & 0.23 & 0.89 \\
\cmidrule(lr){2-9}
    &             & Dyspnea     & 0.63 & 0.95  & 0.54 & 0.95  & 0.44 & 0.92 \\
    & Respiratory & Orthopnea   & 0.29 & 0.67  & 0.31 & 0.66  & 0.33 & 0.87 \\
    &             & Respiratory & 0.35 & 0.64  & 0.40 & 0.48  & 0.19 & 0.31 \\    
\cmidrule(lr){2-9}
    & Composite   & Composite   & 4.66 & 0.61  & 4.66 & 0.61  & 3.92 & 0.68 \\
\midrule
\textbf{P3}  &             & Speech      & 0.17 & 0.26  & 0.18 & 0.26  & 0.21 & -0.41 \\
    & Bulbar      & Salivation  & 0.36 & 0.00  & 0.36 & 0.00  & 0.31 & 0.08 \\
    &             & Swallowing  & 0.15 & 0.00  & 0.15 & 0.00  & 0.17 & 0.00 \\
\cmidrule(lr){2-9}
    &             & Handwriting & 0.11 & 0.00  & 0.11 & 0.00  & 0.08 & 0.00 \\
    & Fine Motor  & Cutting     & 0.03 & 0.00  & 0.03 & 0.00  & {--} & {--} \\
    &             & Dressing    & 0.16 & 0.09  & 0.16 & 0.23  & 0.12 & 0.79 \\
\cmidrule(lr){2-9}
    &             & Turning     & 0.19 & 0.00  & 0.18 & 0.00  & 0.29 & 0.32 \\
    & Gross Motor & Walking     & 0.14 & -0.32 & 0.14 & -0.37 & 0.20 & -0.43 \\
    &             & Stairs      & 0.26 & 0.00  & 0.26 & 0.00  & 0.06 & 0.00 \\
\cmidrule(lr){2-9}
    &             & Dyspnea     & 0.38 & 0.63  & 0.39 & 0.59  & 0.32 & 0.18 \\
    & Respiratory & Orthopnea   & 0.32 & 0.95  & 0.24 & 0.96  & 0.11 & 0.00 \\
    &             & Respiratory & 0.09 & 0.76  & 0.10 & 0.77  & 0.16 & 0.83 \\    
\cmidrule(lr){2-9}
    & Composite   & Composite   & 2.27 & -0.69 & 2.21 & -0.63 & 4.44 & 0.08 \\
\bottomrule
\end{tabular*}
\end{table}